\documentclass[twoside,journal]{IEEEtran}
\usepackage{amsmath,amsfonts}
\usepackage{algorithmic}
\usepackage{algorithm}
\usepackage{array}
\usepackage{textcomp}
\usepackage{stfloats}
\usepackage{url}
\usepackage{verbatim}
\usepackage{graphicx}
\usepackage{cite}
\usepackage{booktabs}
\usepackage{colortbl}
\usepackage{xcolor}
\usepackage{subfigure}
\usepackage{rotating}
\usepackage{multirow}
\usepackage{url}
\usepackage{pifont}
\usepackage[hidelinks]{hyperref}
\definecolor{red}{rgb}{0,0,0}
\definecolor{revred}{RGB}{220,0,0}

\begin{document}

\title{HFP-SAM: Hierarchical Frequency Prompted SAM for Efficient Marine Animal Segmentation}
\author{Pingping~Zhang,
        Tianyu Yan,
        Yuhao Wang,
        Yang Liu,
        Tongdan Tang,
        Yili Ma,
        Long Lv,
        Feng~Tian,
        Weibing Sun,
        and~Huchuan~Lu

\thanks{
Copyright (c) 2026 IEEE. Personal use of this material is permitted. However, permission to use this material for any other purposes must  be obtained from the IEEE by sending an email to \textcolor{blue}{\underline{pubs-permissions@ieee.org}}.

Corresponding author: Yang Liu.

Pingping~Zhang, Tianyu Yan, Yuhao Wang and Yang Liu are with the School of Future Technology, Dalian University of Technology. (Email: zhpp@dlut.edu.cn; 2981431354@mail.dlut.edu.cn; 924973292@mail.dlut.edu.cn; ly@dlut.edu.cn)

Tongdan Tang and Yili Ma are with the Central Hospital of Dalian University of Technology. (Email: tangtongdan2002@sina.com; mayili73@163.com)

Long Lv, Feng Tian and Weibing Sun are with the Affiliated Zhongshan Hospital of Dalian University. (Email: lvlong113@126.com; tianfeng73@163.com; massurm@163.com)

Huchuan~Lu is with the School of Information and Communication Engineering, Dalian University of Technology. (Email: lhchuan@dlut.edu.cn)
}
}
\markboth{IEEE Transactions on Image Processing}
{Yan \MakeLowercase{\textit{et al.}}:HFP-SAM: Hierarchical Frequency Prompted SAM for Efficient Marine Animal Segmentation}
\maketitle
\begin{abstract}
Marine Animal Segmentation (MAS) aims at identifying and segmenting marine animals from complex marine environments.
Most of previous deep learning-based MAS methods struggle with the long-distance modeling issue.
Recently, Segment Anything Model (SAM) has gained popularity in general image segmentation.
However, it lacks of perceiving fine-grained details and frequency information.
To this end, we propose a novel learning framework, named Hierarchical Frequency Prompted SAM (HFP-SAM) for high-performance MAS.
First, we design a Frequency Guided Adapter (FGA) to efficiently inject marine scene information into the frozen SAM backbone through frequency domain prior masks.
Additionally, we introduce a Frequency-aware Point Selection (FPS) to generate highlighted regions through frequency analysis.
These regions are combined with the coarse predictions of SAM to generate point prompts and integrate into SAM's decoder for fine predictions.
Finally, to obtain comprehensive segmentation masks, we introduce a Full-View Mamba (FVM) to efficiently extract spatial and channel contextual information with linear computational complexity.
Extensive experiments on four public datasets demonstrate the superior performance of our approach.
The source code is publicly available at https://github.com/Drchip61/TIP-HFP-SAM.
\end{abstract}
\section{Introduction}
\IEEEPARstart{M}arine Animal Segmentation (MAS) aims to accurately identify and segment animals from marine environment.
It is a crucial vision task for advancing marine biology research, ecological monitoring and autonomous underwater robotics.
However, the marine environment poses unique challenges for animal segmentation, including poor visibility, variable lighting conditions, and the presence of particulates.
These factors often obscure marine animal features, making accurate segmentation a challenging task.
Consequently, traditional methods that rely on handcrafted features and energy-based models struggle to perform adequately in this task~\cite{priyadarshni2020underwater}.
The limitations of these approaches highlight the need for advanced segmentation techniques.
\begin{figure}[t]
\centering
\resizebox{0.48\textwidth}{!}
{
\begin{tabular}{@{}c@{}c@{}}
\includegraphics[scale=0.6]{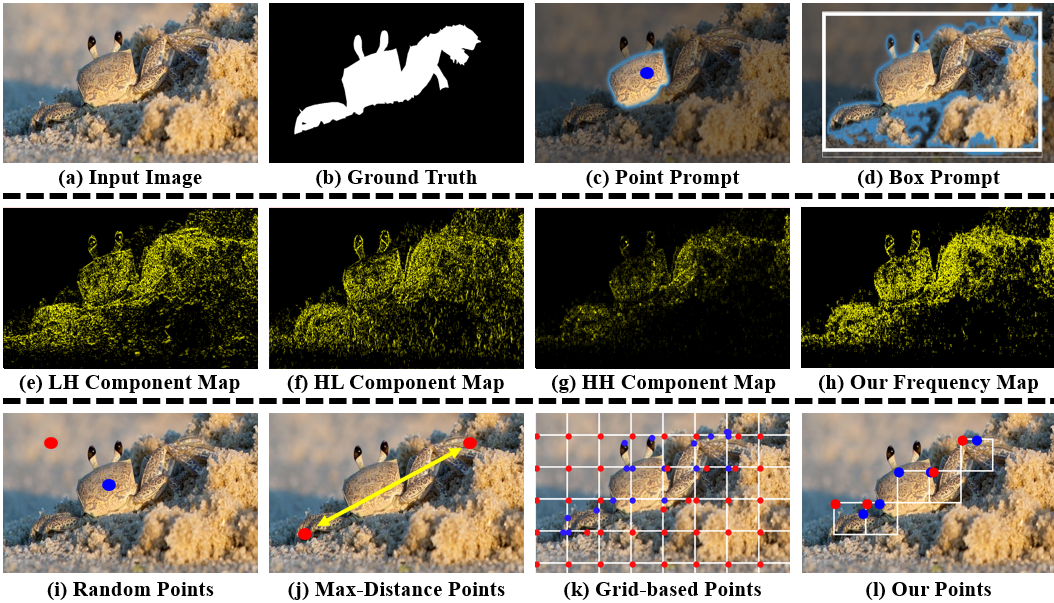} \\
\end{tabular}
}
\caption{Our motivations and advantages. The first row shows the input image, ground truth, the prediction mask obtained by point prompt and box prompt, respectively.
The second row shows the component maps obtained from the wavelet transform and the result of our combined frequency map.
The third row displays the existing point prompt methods and our proposed method.
}
\label{motivation}
\end{figure}

Deep learning has significantly advanced image segmentation, with Convolutional Neural Network (CNN) based methods excelling in local information extraction~\cite{long2015fully,he2016deep,huang2017densely} and unsupervised hierarchical representation~\cite{li2023x,li2024model,li2025enhanced}.
However, their reliance on local receptive fields limits their ability to capture global context, which is crucial for distinguishing marine entities from their environments.
Recently, Transformer~\cite{dosovitskiy2020image} has demonstrated its strength in modeling long-range dependencies, particularly in vision tasks requiring a global understanding.
However, Transformers typically demand large-scale datasets for training \cite{dosovitskiy2020image,touvron2021training,steiner2021train}, which poses a challenge for vision tasks with limited data, such as MAS.
Recently, Segment Anything Model (SAM)~\cite{kirillov2023segment} emerges as a powerful tool in general image segmentation, trained on a large-scale dataset with one billion samples.
Despite its impressive performance, SAM struggles to specific domains.
Currently, there are some works that have adapted SAM to the MAS task~\cite{zhang2024fantastic,yan2024mas}.
However, these works mainly focus on modifying the encoder and decoder parts of SAM to enhance feature representation abilities.
They pay less attention to the prompt of SAM, making it challenging to achieve further performance improvements.
In fact, SAM needs carefully-designed prompts to achieve better image segmentation.
Motivated by this fact, we emphasize prompt designs for SAM and introduce more reliable prompts in noisy marine scenes.
Fig.~\ref{motivation} (c) and (d) illustrate the limitations of simple point prompts and box prompts.
Generally, point prompts have limitations for segmenting marine animals with complex structures.
In contrast, box prompts often include background content, which can have adverse effects on marine animals that closely resemble the background.
Additionally, marine scene images contain abundant high-frequency noise, making SAM highly susceptible to noise interference despite its strong ability to learn spatial domain information.
This susceptibility can lead to artifacts in the segmentation results.
Furthermore, SAM's simplistic decoder often leads to the loss of detail information.

To address aforementioned issues, we propose a novel learning framework named HFP-SAM for high performance MAS.
As shown in the second row of Fig.~\ref{motivation}, the wavelet-transformed frequency spectrum effectively reduces noise in the marine scene.
By utilizing frequency domain priors, we first introduce a Frequency Guided Adapter (FGA) to inject underwater scene information into the segmentation process.
Then, we design a Frequency-aware Point Selection (FPS) to generate effective point prompts by combining frequency domain priors with the coarse segmentation mask from SAM.
To obtain comprehensive segmentation masks, we design a Full View Mamba (FVM), which possesses excellent long-distance modeling abilities while maintaining linear computation complexity.
Extensive experiments show that our method achieves outstanding performances on four MAS datasets.

In summary, our contributions are as follows:
\begin{itemize}
\item We introduce a novel framework named HFP-SAM that fully leverages frequency domain prior knowledge for marine animal segmentation.
\item We propose a Frequency-aware Point Selection (FPS), which is capable of identifying effective point prompts via frequency domain prior knowledge.
\item We propose a Full View Mamba (FVM), which fully models the comprehensive relationship of deep features and maintains linear computational complexity.
\item We conduct extensive experiments to validate the effectiveness of the whole framework. Our method achieves state-of-the-art performances on four MAS datasets.
\end{itemize}
\section{Related Work}
\label{sec:formatting}
\subsection{Marine Animal Segmentation}
The Marine Animal Segmentation (MAS) task encourages unique challenges, due to the complex and dynamic nature of marine environments.
Very early, methods~\cite{bay2008speeded,lowe2004distinctive,priyadharsini2019object} with handcrafted features prove ineffective due to the complex marine scenes.
Additionally, energy-based methods~\cite{lane1998robust,priyadarshni2020underwater,shihavuddin2013automated} rely on the distribution of image intensities or colors.
They struggle to capture the high-level semantic information in complex marine scenes.

The advent of deep learning methods marks a significant advancement in MAS.
Various methods~\cite{burguera2020segmentation,islam2020simultaneous,king2018comparison} leverage the power of deep features and offer better accuracy and adaptability.
For example, Li \emph{et al.}~\cite{li2021marine} introduce an enhanced cascade decoder network to segment marine animals within challenging underwater environments.
Chen \emph{et al.}~\cite{chen2022robust} employ a siamese network to harness multi-level features and enhance the utilization of global context information.
Cheng \emph{et al.}~\cite{cheng2023bidirectional} integrate local and global cues to improve object segmentation from camouflaged backgrounds.
Fu \emph{et al.}~\cite{fu2023masnet} employ a siamese fusion-based deep neural network to learn semantic representations of marine animals.
However, the local receptive fields limit their ability to model long-distance relationships within marine images.

Recently, Vision Transformer~\cite{dosovitskiy2020image} demonstrates great potential in extracting long-range dependencies and global contexts, resulting in significant improvements for various image segmentation tasks~\cite{zheng2021rethinking,ranftl2021vision,liu2021tritransnet}.
Especially, Hong \emph{et al.}~\cite{hong2023usod10k} introduce Transformers into MAS and achieve improvements.
However, a notable hurdle for Transformers is their reliance on large-scale training datasets~\cite{dosovitskiy2020image}.
Presently, the marine domain lacks such large-scale datasets for training Transformers.
In addition, Zhang \emph{et al.}~\cite{zhang2024fantastic} firstly introduce SAM into MAS and achieve an obvious improvement compared with previous Transformer-based methods.
Yan \emph{et al.}~\cite{yan2024mas} integrate multi-level feature maps from SAM's encoder by a novel hypermap design.
However, they only focus on the SAM's encoder and decoder, but neglect the importance of SAM's prompt encoder.
\begin{figure*}
\centering
\resizebox{1.0\textwidth}{!}
{
\begin{tabular}{@{}c@{}c@{}}
\includegraphics[scale=1]{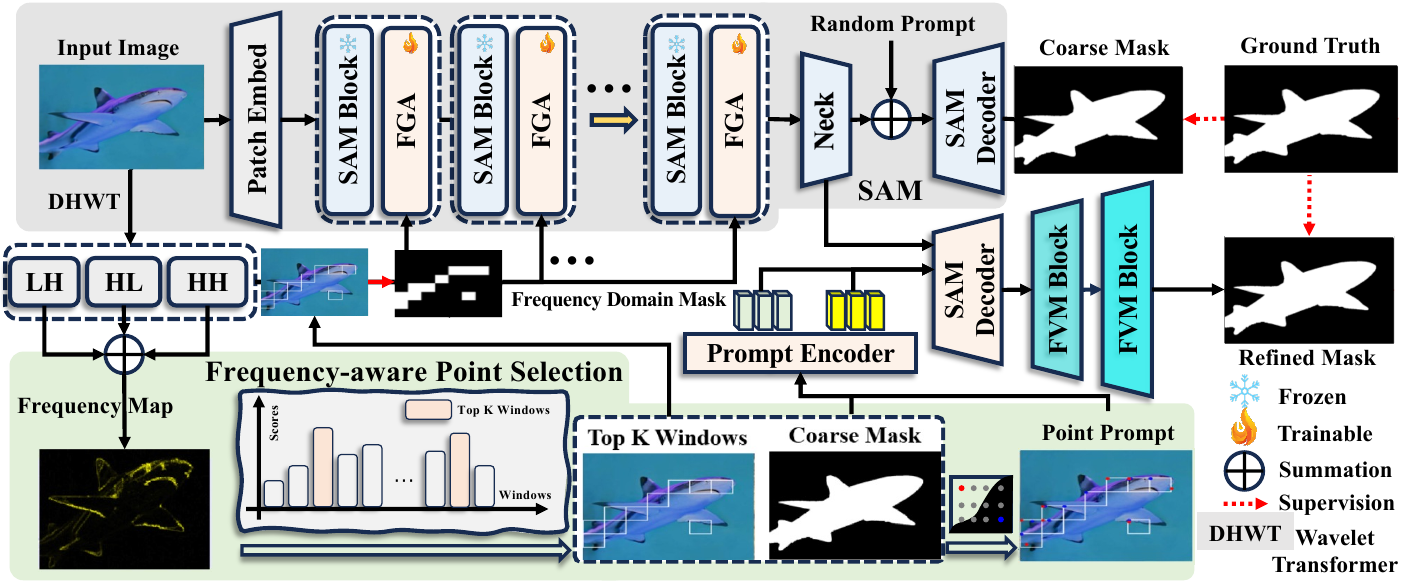} \\
\end{tabular}
}
\caption{The framework of our proposed HFP-SAM. It consists of three main components: Frequency Guided Adapter (FGA), Frequency-aware Point Selection (FPS), and Full View Mamba (FVM).
HFP-SAM leverages frequency domain information to automatically obtain efficient prompts.
Additionally, HFP-SAM utilizes the FVM to fully combine frequency and spatial domain information.
}
\label{frame}
\end{figure*}
\subsection{Segment Anything Model for Customized Tasks}
Recently, SAM~\cite{kirillov2023segment} represents a significant leap forward in general image segmentation.
However, adaptations of SAM have revealed limitations for customized tasks.
Typically, the adapters~\cite{zhang2023customized,wang2023sam,gong20233dsam,chen2023sam} within SAM mainly comprise basic linear projection layers, neglecting frequency domain information.
To address this issue, Chen \emph{et al.}~\cite{wei2024medsam} propose a frequency domain adapter.
However, direct frequency-domain encoding struggles to align with the spatial representations learned during SAM's pre-training.
Motivated by this, we leverage frequency prior masks as guidance to modulate pre-trained features. It helps narrow the alignment gap between frequency cues and spatial representations learned during SAM pre-training.
Furthermore, SAM's prompts are divided into two main types: handcrafted prompt and automatic prompt.
The weakness with handcrafted prompt methods~\cite{shen2023temporally,dai2023samaug} lies in their unsuitability for fully automated segmentation tasks.
On the other hand, automatic prompt methods~\cite{chen2024rsprompter,jie2023adaptershadow} introduce additional modules to generate point prompts, which will bring more computational overhead.
To address these problems, Dai \emph{et al.}~\cite{dai2023samaug} generate point prompts by randomly selecting points or selecting points at the maximum distance.
However, the effectiveness of points obtained through this method remains insufficient.
Thus, there is a great need of a lightweight yet reliable automatic prompt scheme.
In this work, we obtain efficient and critical point prompts by leveraging frequency domain information and the coarse segmentation mask, avoiding heuristic sampling and extra prompt-generation networks.

Additionally, the simplicity of SAM's decoder often results in the loss of crucial detail information.
Some efforts~\cite{xiong2024sam2,alzate2023sam} have attempted to implement layer-by-layer up-sampling decoders.
However, the inherent limitations of convolutions have hampered the performance.
Meanwhile, employing Transformers leads to more computational overload.
To address these challenges, we incorporate the Mamba structure~\cite{gu2023mamba}, which maintains a global long-range modeling while ensuring linear complexity.
Some works~\cite{liu2024vmamba,zhu2024vision,guo2024mambair} have attempted to seamlessly integrate the Mamba architecture for vision tasks.
In this paper, we fully unleash the potential of Mamba through a comprehensive perspective.

In summary, our proposed key components are very different from existing SAM adaptation methods, frequency-domain modules, prompt generation strategies, and Mamba-based decoders.
In brief, our FGA uses frequency prior masks to guide feature modulation (instead of directly encoding frequency features).
Our FPS generates boundary-aware point prompts by combining frequency cues with SAM's coarse mask without extra prompt-generation networks.
Our FVM further models channel dependencies via bidirectional channel SSM in addition to spatial scanning.
\section{Proposed Method}
As shown in Fig.~\ref{frame}, our method includes three main components: Frequency Guided Adapter (FGA), Frequency-aware Point Selection (FPS) and Full-View Mamba (FVM).
First, we obtain a coarse segmentation mask using SAM.
Then, we use the FPS to generate effective point prompts by leveraging frequency domain information along with the coarse segmentation mask.
The points generated by FPS, along with the coarse mask, are fed as prompts into SAM's prompt encoder.
Finally, we obtain a refined segmentation mask through the FVM.
These key components are elaborated in the following sections.
\begin{figure}[h]
\centering
\resizebox{0.4\textwidth}{!}
{
\begin{tabular}{@{}c@{}c@{}}
\includegraphics[scale=0.99]{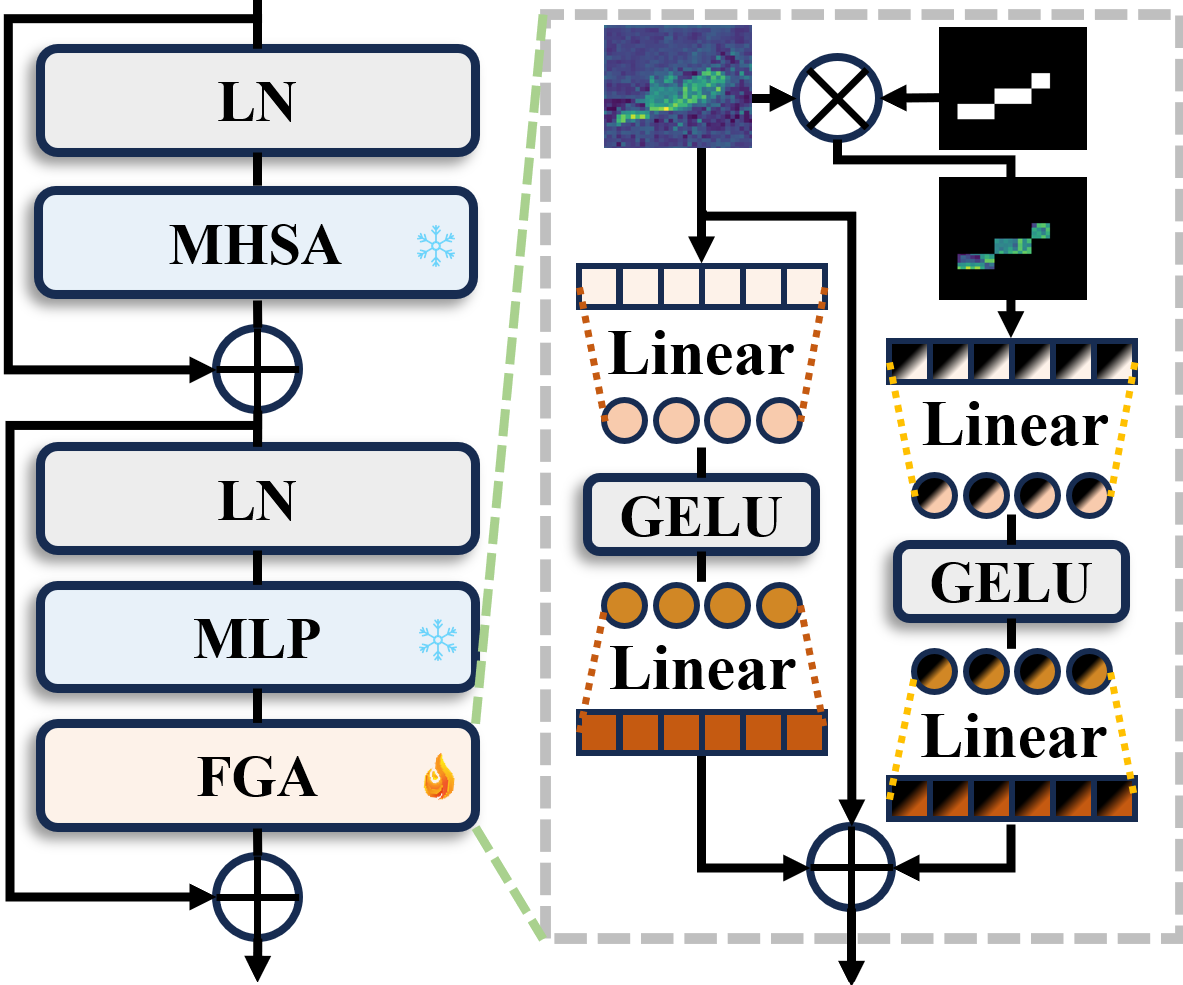} \\
\end{tabular}
}
\caption{Illustration of the Frequency Guided Adapter.
}
\label{FGA}
\end{figure}
\subsection{Frequency Guided Adapter}
The original pre-trained backbone of SAM possesses strong capabilities in image understanding but struggles for downstream tasks, especially in MAS.
As a result, many works\cite{chen2023sam,lai2023detect} use adapters to inject specific domain information into the frozen SAM for customized tasks.
However, previous adapters focus solely on spatial information.
Spatial domain information is highly sensitive to high-frequency noise. In marine scenes, relying solely on spatial information can easily lead to noise interference.
Consequently, we introduce Frequency Guided Adapter (FGA) to inject marine domain information with frequency domain prior knowledge.
As shown in Fig.~\ref{FGA}, we use frequency domain prior masks as guidance to combine with the feature maps.
More specifically, $X_{i}\in \mathbb{R} ^{N\times D} $ is the input of the $i$-th Transformer block, where $N$ represents the number of tokens, and $D$ represents the embedding dimension.
The procedure can be represented as follows:
\begin{equation}
\overline{X} _{i} =\mathrm{MHSA}(\mathrm{LN}(X_{i}))+X_{i},
\end{equation}
\begin{equation}
\widehat{X}  _{i} =\mathrm{MLP}(\mathrm{LN}(\overline{X} _{i})),
\end{equation}
where \(\mathrm{LN}(\cdot)\) refers to the Layer Normalization~\cite{ba2016layer}, MHSA stands for the multi-head self-attention, and MLP denotes the multi-layer perceptron.
Then, we make use of frequency information to process the feature map $\widehat{X}_i$.
Technically, we first use the Discrete Haar Wavelet Transform (DHWT)~\cite{mallat1989theory} to decompose the input image $I$ into four sub-bands:
\begin{equation}
I^{ll},I^{lh},I^{hl},I^{hh} = \mathrm{DHWT}(I).
\end{equation}
Here, $I^{ll}$ is the low-frequency sub-band. $I^{lh}$, $I^{hl}$, and $I^{hh}$ are the horizontal, vertical and diagonal high-frequency sub-bands, respectively.
Then, we average the three high-frequency sub-bands to obtain our frequency map $M^{h}$:
\begin{equation}
    M^{h} = \frac{I^{lh}+I^{hl}+I^{hh}}{3}.
    \label{mh}
\end{equation}
In $M^{h}$, regions with high response values highlight significant changes in the image, including the edges between the target and the background.
Then, we use a sliding window $S^{w}$ to calculate the average frequency variation value of each window in $M^{h}$.
After that, we select windows with the top $k$ response values as the prior regions $P$:
\begin{equation}
    P = \mathrm{Top}_{k}^{w}(S^{w}(M^{h})).
    \label{topw}
\end{equation}
To ensure the same size with the feature map, we down-sample the frequency domain prior mask to match the dimensions.
Then, we multiply it with the feature map, producing a frequency-guided feature map:
\begin{equation}
\widehat{X}_{i}^{f} = \widehat{X}_{i}\odot \theta(P),
\end{equation}
where $\theta$ is the down-sample operation and $\widehat{X}_{i}^{f}\in \mathbb{R}^{\frac{N}{16\times16} \times D}$ is the frequency guided feature map.
Ultimately, we inject the feature maps $\widehat{X}_{i}$ and $\widehat{X}_{i}^{f}$ into SAM's backbone.
The operation is represented as follows:
\begin{equation}
\begin{split}
    X_{i+1}= W_{f}^{Up}(\phi (W_{f}^{Down} (\widehat{X}_{i}^{f})))\\
    +W_{s}^{Up}(\phi (W_{s}^{Down} (\widehat{X}_{i})))+\widehat{X}_{i}.
\end{split}
\end{equation}
where $\phi$ is the Gaussian Error Linear Unit (GELU)~\cite{hendrycks2016gaussian}.
$W_{s/f}^{Down}\in \mathbb{R}^{D\times r} $ and $W_{s/f}^{Up}\in \mathbb{R}^{r\times D} $ represent the weights of two linear projections aimed at decreasing and subsequently increasing the feature dimension, respectively.
$r$ denotes the reduction ratio.
FGA facilitates the adaptation of SAM for marine scene segmentation via frequency priors.
\subsection{Frequency-aware Point Selection}
Accurate and detailed prompts can significantly enhance SAM's segmentation performance.
However, the high level of noise in marine scenes makes it challenging to obtain high-quality prompts using only spatial information.
Moreover, marine animals have complex shapes, especially with highly irregular edges.
This reduces the effectiveness of conventional point prompts.
Therefore, we use frequency domain prior information to significantly filter out noise interference, allowing the model to focus more on the edges of marine animals.
Following Eq.~\ref{mh}, in each selected window, we sample point prompts with the coarse segmentation results $M^{c}$.
Specifically, we select ${2t}$ point prompts within each selected window, named $p^{2t}$, including $t$ highest and $t$ lowest values.
Since $M^{c}$ is a probability map, we can use a threshold $\tau$ to obtain the binary segmentation results $M^{b}$:
\begin{equation}
M^{b} = \left\{
\begin{aligned}
1, & \quad M^{c} > \tau, \\
0, & \quad M^{c} \leq \tau,
\end{aligned}
\right.
\end{equation}
Then, we can generate the final point prompts.
The sampling process can be represented as follows:
\begin{equation}
\begin{split}
p^{+} = \{p|M^{b}(p^{2t}) = 1\}, p^{-} = \{p|M^{b}(p^{2t}) = 0\},
\end{split}
\end{equation}
where $p^{+}$ and $p^{-}$ are the positive point prompts and the negative point prompts, respectively.
Finally, we send the point prompts $p^{+}$ and $p^{-}$ to SAM and obtain the final segmentation result $M^{f}$.
Through the FPS, we can obtain efficient point prompts without external input, enabling SAM to focus more on regions with significant frequency domain changes.
\begin{figure}
\centering
\resizebox{0.4\textwidth}{!}
{
\begin{tabular}{@{}c@{}c@{}}
\includegraphics[scale=0.9]{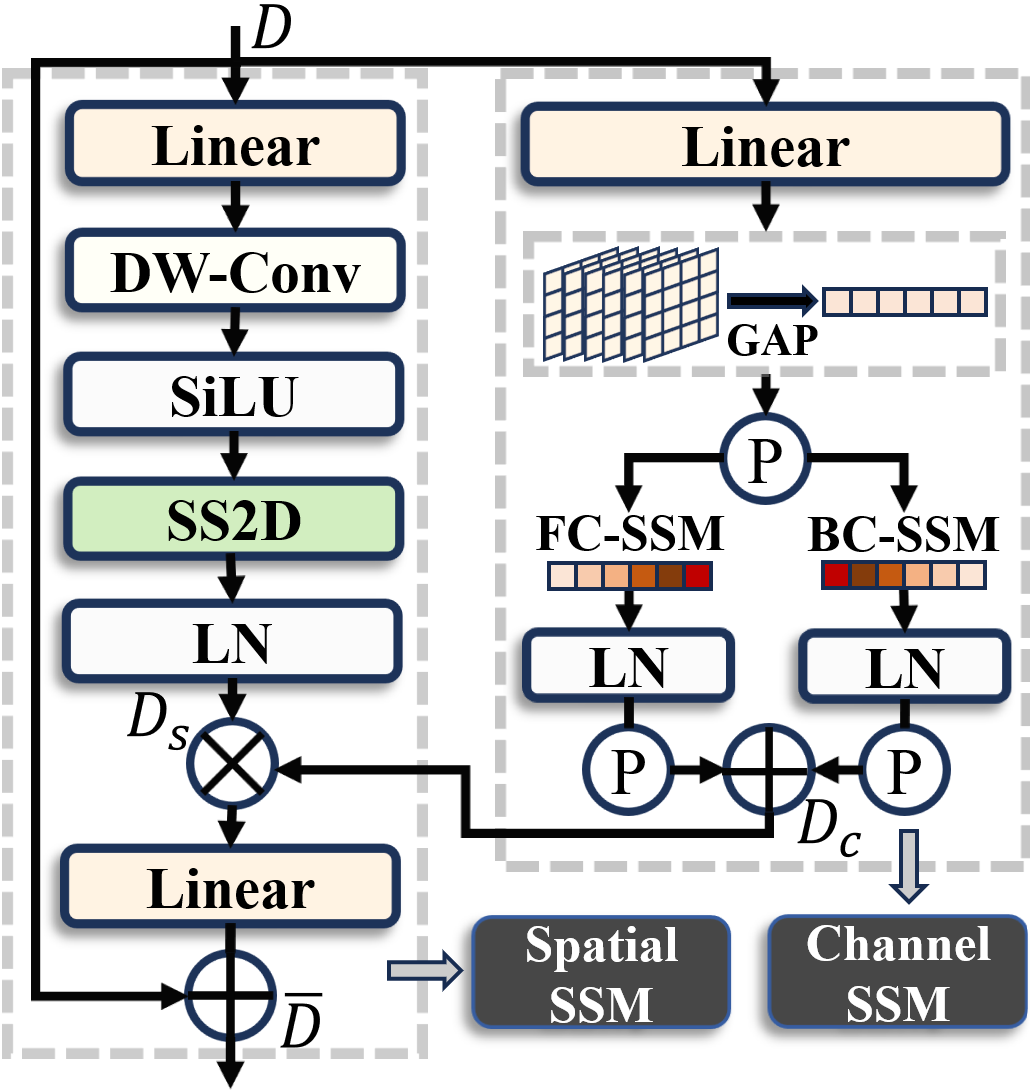} \\
\end{tabular}
}
\caption{Illustration of the Full View Mamba.
}
\label{fvm}
\end{figure}
\subsection{Full View Mamba}
After passing through the FGA and FPS, the feature map can contain both spatial and frequency domain information.
Then, we propose a Full View Mamba (FVM), which comprehensively processes the feature map from both channel and spatial views.
By scanning along the channel dimension, FVM captures global inter-channel correlations.
Meanwhile, scanning along the spatial dimension provides long-range spatial modeling to propagate contextual information across the whole feature map.
In this way, we can achieve a more full integration of spatial and frequency domain features.
Besides, we adopt the State Space Model (SSM)~\cite{liu2024vmamba} as a linear-time alternative to self-attention for long-range modeling, enabling efficient global context aggregation in the decoder.
\begin{table*}[h]
\renewcommand\arraystretch{1.1}
\setlength\tabcolsep{5.5pt}
\centering
\caption{Performance comparisons on MAS3K and RMAS. Each group represents methods based on different architectures.}
\resizebox{0.86\textwidth}{!}
{
\begin{tabular}{c|c|c|c|c|c|c|c|c|c|c|c}
\hline
& &\multicolumn{5}{c|}{\textbf{MAS3K}}        &\multicolumn{5}{c}{\textbf{RMAS}} \\ \cline{3-12}
\textbf{Method}&\textbf{Backbone}&\textbf{mIoU} & $\textbf{S}_\alpha$ & $\textbf{F}_\beta^w $&$ \textbf{mE}_\phi$ & \textbf{MAE}& \textbf{mIoU} & $\textbf{S}_\alpha$ & $\textbf{F}_\beta^w$ & $\textbf{mE}_\phi$ & \textbf{MAE} \\
\hline
 UNet++~\cite{zhou2018unet++}   &--&0.506&0.726&0.552&0.790&0.083&0.558&0.763&0.644&0.835&0.046\\
 BASNet~\cite{qin2019basnet}   &ResNet-34&0.677&0.826&0.724&0.862&0.046&0.707&0.847&0.771&0.907&0.032\\
 PFANet~\cite{zhao2019pyramid}   &VGG-16&0.405&0.690&0.471&0.768&0.086&0.556&0.767&0.582&0.810&0.051\\
 SCRN~\cite{wu2019stacked}   &ResNet-50&0.693&0.839&0.730&0.869&0.041&0.695&0.842&0.731&0.878&0.030\\
 U2Net~\cite{qin2020u2}   &--&0.654&0.812&0.711&0.851&0.047&0.676&0.830&0.762&0.904&0.029\\
 SINet~\cite{fan2020camouflaged} &ResNet-50&0.658&0.820&0.725&0.884&0.039&0.684&0.835&0.780&0.908&0.025\\
 PFNet~\cite{mei2021camouflaged}  &ResNet-50&0.695&0.839&0.746&0.890&0.039&0.694&0.843&0.771&0.922&0.026\\
 RankNet~\cite{lv2021simultaneously} &ResNet-50&0.658&0.812&0.722&0.867&0.043&0.704&0.846&0.772&0.927&0.026\\
 C2FNet~\cite{sun2021context}   &Res2Net-50&0.717&0.851&0.761&0.894&0.038&0.721&0.858&0.788&0.923&0.026\\
 ECDNet~\cite{li2021marine}   &ResNet-50&0.711&0.850&0.766&0.901&0.036&0.664&0.823&0.689&0.854&0.036\\
 OCENet~\cite{liu2022modeling}  &ResNet-50&0.667&0.824&0.703&0.868&0.052&0.680&0.836&0.752&0.900&0.030\\
 ZoomNet~\cite{pang2022zoom}   &ResNet-50&0.736&0.862&0.780&0.898&0.032&0.728&0.855&0.795&0.915&0.022\\
 MASNet~\cite{fu2023masnet} &Res2Net-50&0.742&0.864&0.788&0.906&0.032&0.731&0.862&0.801&0.920&0.024\\
 \hline
 SETR~\cite{zheng2021rethinking}   &ViT-L&0.715&0.855&0.789&0.917&0.030&0.654&0.818&0.747&0.933&0.028\\
 TransUNet~\cite{chen2021transunet}   &ResNet-50+ViT-B&0.739&0.861&0.805&0.919&0.029&0.688&0.832&0.776&0.941&0.025\\
 H2Former~\cite{he2023h2former}   &--&0.748&0.865&0.810&0.925&0.028&0.717&0.844&0.799&0.931&0.023\\
 \hline
 SAM~\cite{kirillov2023segment}  &ViT-B&0.566&0.763&0.656&0.807&0.059&0.445&0.697&0.534&0.790&0.053\\
 I-MedSAM\cite{wei2024medsam}   &ViT-B&0.698&0.835&0.759&0.889&0.039&0.633&0.803&0.699&0.893&0.035\\
 Med-SAM\cite{wu2023medical}   &ViT-B&0.739& 0.861& 0.811 &0.922& 0.031&0.678&0.832&0.778&0.920&0.027\\
 SAM-Adapter\cite{chen2023sam}   &ViT-H&0.714&0.847&0.782&0.914&0.033&0.656&0.816&0.752&0.927&0.027\\
 SAM-DADF~\cite{lai2023detect}   &ViT-H&0.742&0.866&0.806&0.925&0.028&0.686&0.833&0.780&0.926&0.024\\
 MAS-SAM\cite{yan2024mas}  &ViT-B&0.788&0.887&0.840&0.938&0.025&0.742&0.865&0.819&0.948&0.021\\
 Dual-SAM~\cite{zhang2024fantastic}   &ViT-B&0.789&0.884&0.838&0.933&0.023&0.735&0.860&0.812&0.944&0.022\\
\hline
SAM2\cite{ravi2024sam}&Hiera-L&0.602&0.784&0.705&0.828&0.050&0.395&0.602&0.403&0.623&0.168\\
SAM2-Adapter\cite{chen2024sam2}&Hiera-L&0.778&0.862&0.824&0.923&0.027&0.650&0.791&0.702&0.896&0.036\\
\hline
\textbf{HFP-SAM}  &\textbf{ViT-B}&\textbf{0.797}&\textbf{0.888}&\textbf{0.845}&\textbf{0.938}&\textbf{0.024}&\textbf{0.745}&\textbf{0.865}&\textbf{0.820}&\textbf{0.946}&\textbf{0.021}\\
\textbf{HFP-SAM2}  &\textbf{Hiera-L}&\textbf{0.807}&\textbf{0.891}&\textbf{0.863}&\textbf{0.945}&\textbf{0.022}&\textbf{0.758}&\textbf{0.869}&\textbf{0.833}&\textbf{0.957}&\textbf{0.019}\\
\hline
\end{tabular}
}
\label{mas3k}
\end{table*}
\begin{table*}[h]
\renewcommand\arraystretch{1.1}
\setlength\tabcolsep{5.5pt}
\centering
\caption{Performance comparisons on UFO120 and RUWI. Each group represents methods based on different architectures.}
\resizebox{0.86\textwidth}{!}
{
\begin{tabular}{c|c|c|c|c|c|c|c|c|c|c|c}
\hline
& &\multicolumn{5}{c|}{\textbf{UFO120}}        &\multicolumn{5}{c}{\textbf{RUWI}} \\ \cline{3-12}
\textbf{Method}&\textbf{Backbone}&\textbf{mIoU} & $\textbf{S}_\alpha$ & $\textbf{F}_\beta^w $&$ \textbf{mE}_\phi$ & \textbf{MAE}& \textbf{mIoU} & $\textbf{S}_\alpha$ & $\textbf{F}_\beta^w$ & $\textbf{mE}_\phi$ & \textbf{MAE} \\
\hline
    UNet++~\cite{zhou2018unet++}    &--&0.412&0.459&0.433&0.451&0.409&0.586&0.714&0.678&0.790&0.145\\
    BASNet~\cite{qin2019basnet}   &ResNet-34&0.710&0.809&0.793&0.865&0.097&0.841&0.871&0.895&0.922&0.056\\
    PFANet~\cite{zhao2019pyramid}   &VGG-16&0.677&0.752&0.723&0.815&0.129&0.773&0.765&0.811&0.867&0.096\\
    SCRN~\cite{wu2019stacked}   &ResNet-50&0.678&0.783&0.760&0.839&0.106&0.830&0.847&0.883&0.925&0.059\\
    U2Net~\cite{qin2020u2}   &--&0.680&0.792&0.709&0.811&0.134&0.841&0.873&0.861&0.786&0.074\\
    SINet~\cite{fan2020camouflaged}   &ResNet-50&0.767&0.837&0.834&0.890&0.079&0.785&0.789&0.825&0.872&0.096\\
    PFNet~\cite{mei2021camouflaged}   &ResNet-50&0.570&0.708&0.550&0.683&0.216&0.864&0.883&0.870&0.790&0.062\\
    RankNet~\cite{lv2021simultaneously}   &ResNet-50&0.739&0.823&0.772&0.828&0.101&0.865&0.886&0.889&0.759&0.056\\
    C2FNet~\cite{sun2021context}   &Res2Net-50&0.747&0.826&0.806&0.878&0.083&0.840&0.830&0.883&0.924&0.060\\
    ECDNet~\cite{li2021marine}   &ResNet-50&0.693&0.783&0.768&0.848&0.103&0.829&0.812&0.871&0.917&0.064\\
    OCENet~\cite{liu2022modeling}  &ResNet-50&0.605&0.725&0.668&0.773&0.161&0.763&0.791&0.798&0.863&0.115\\
    ZoomNet~\cite{pang2022zoom}   &ResNet-50&0.616&0.702&0.670&0.815&0.174&0.739&0.753&0.771&0.817&0.137\\
    MASNet~\cite{fu2023masnet} &Res2Net-50&0.754&0.827&0.820&0.879&0.083&0.865&0.880&0.913&0.944&0.047\\
    \hline
    SETR~\cite{zheng2021rethinking}   &ViT-L&0.711&0.811&0.796&0.871&0.089&0.832&0.864&0.895&0.924&0.055\\
    TransUNet~\cite{chen2021transunet}   &ResNet-50+ViT-B&0.752&0.825&0.827&0.888&0.079&0.854&0.872&0.910&0.940&0.048\\
    H2Former~\cite{he2023h2former}   &--&0.780&0.844&0.845&0.901&0.070&0.871&0.884&0.919&0.945&0.045\\
    \hline
    SAM~\cite{kirillov2023segment}   &ViT-B&0.681&0.768&0.745&0.827&0.121&0.849&0.855&0.907&0.929&0.057\\
    I-MedSAM\cite{wei2024medsam}   &ViT-B&0.730&0.818&0.788&0.865&0.084&0.844&0.849&0.897&0.923&0.050\\
    Med-SAM\cite{wu2023medical}   &ViT-B&0.774& 0.842& 0.839 &0.899& 0.072&0.877&0.885&0.921&0.942&0.045\\
    SAM-Adapter~\cite{chen2023sam}   &ViT-H&0.757&0.829&0.834&0.884&0.081&0.867&0.878&0.913&0.946&0.046\\
    SAM-DADF~\cite{lai2023detect}   &ViT-H&0.768&0.841&0.836&0.893&0.073&0.881&0.889&0.925&0.940&0.044\\
    MAS-SAM\cite{yan2024mas}  &ViT-B&0.807&0.861&0.864&0.914&0.063&0.902&0.894&0.941&0.961&0.035\\
    Dual-SAM~\cite{zhang2024fantastic}   &ViT-B&0.810&0.856&0.864&0.914&0.064&0.900&0.903&0.935&0.959&0.035\\
   \hline
   SAM2\cite{ravi2024sam}&Hiera-L&0.692&0.760&0.725&0.805&0.135&0.825&0.806&0.859&0.907&0.716\\
   SAM2-Adapter\cite{chen2024sam2}&Hiera-L&0.755& 0.803& 0.797 &0.853& 0.095&0.883&0.891&0.925&0.947&0.042\\
\hline
\textbf{HFP-SAM}   &\textbf{ViT-B}&\textbf{0.803}&\textbf{0.865}&\textbf{0.862}&\textbf{0.918}&\textbf{0.061}&\textbf{0.904}&\textbf{0.893}&\textbf{0.936}&\textbf{0.964}&\textbf{0.034}\\
\textbf{HFP-SAM2}&\textbf{Hiera-L}&\textbf{0.813}&\textbf{0.870}&\textbf{0.868}&\textbf{0.921}&\textbf{0.058}&\textbf{0.913}&\textbf{0.906}&\textbf{0.951}&\textbf{0.969}&\textbf{0.031}\\
\hline
\end{tabular}
}
\label{ufo}
\end{table*}

To be specific, our FVM is composed of spatial SSM and channel SSM.
Compared with existing SSM variants, our FVM additionally models channel dependencies and fuses them with spatial features, forming a full-view decoder block.
Spatial SSM integrates the spatial information of input sequences with four scanning directions.
Channel SSM across the channel dimension will be scanned in two directions: forward and backward.
In Fig. \ref{fvm}, we first obtain the outcome $D \in \mathbb{R}^{\frac{H}{4}\times\frac{W}{4} \times C}$ from SAM's decoder.
Then, $D$ is fed into the spatial SSM and the channel SSM to obtain the spatial and channel-wise features.
In the spatial SSM, the operations can be represented as follows:
\begin{equation}
D_{s} = \mathrm{LN}(\mathrm{SS2D}(\psi(\mathrm{DWC}(\mathrm{Linear}(D))))))
\end{equation}
where $\mathrm{Linear}$ is a linear transformation, $\mathrm{DWC}$ is a depth-wise convolution, $\psi$ is the Sigmoid Linear Unit (SiLU)~\cite{hendrycks2016gaussian}, and $\mathrm{SS2D}$ is the 2D spatial scanning strategy.
For channel SSM, we first perform the following operations:
\begin{equation}
  F_{c} = \mathcal{P}(\mathrm{GAP}(\mathrm{Linear}(D))),
\end{equation}
where $\mathrm{GAP}$ is the global average pooling, $\mathcal{P}$ is the matrix transpose operation, and $F_{c}$ is obtained through $\mathcal{P}$ to form the spatial sequences as before.
After that, $F_{c}$ undergoes the bidirectional SSM to model the channel-wise relations:
\begin{equation}
D_{c} =\mathcal{P}(\mathrm{LN}(\overrightarrow{\mathrm{SSM}}(F_{c}))) + \mathcal{P}(\mathrm{LN}(\overleftarrow{\mathrm{SSM}}(F_{c}))),
\end{equation}
where $\overrightarrow{\mathrm{SSM}}$ and $\overleftarrow{\mathrm{SSM}}$ are the Forward Channel-wise SSM (FC-SSM) and Backward Channel-wise SSM (BC-SSM), respectively.
Then, the spatial and channel-wise features are multiplied to obtain the fused feature:
\begin{equation}
D^{*} = D_{s} \odot D_{c}.
\end{equation}
After the linear reduction and residual connection, the final feature can be obtained as:
\begin{equation}
\overline{D} = \mathrm{Linear}(D^{*}) + D.
\end{equation}
Through the above process, the spatial and channel-wise features are effectively captured and integrated.
\subsection{Loss Function}
To train our framework, we adopt a weighted binary cross-entropy loss and a weighted intersection over union loss.
This combined loss is designed to enhance the boundary preservation and focus on challenging regions.
More specifically, the weighted binary cross-entropy loss is defined as:
\begin{equation}
    \mathcal{L}_{\text{wbce}} = \sum w \cdot {\text{CE}}(M^{f}, M^{gt})
\end{equation}
where $M^{f} \in \mathbb{R}^{H \times W}$ is the predicted mask, and $M^{gt} \in \{0, 1\}^{H \times W}$ is the ground truth mask.
$\text{CE}(\cdot)$ is the element-wise cross-entropy.
The weight map $w$ is defined as the previous work~\cite{fan2020pranet}.
This weighted strategy gives a higher importance to boundary regions, encouraging the model to learn more precise object contours.

In addition, we compute the weighted intersection over union loss as follows:
\begin{equation}
    \mathcal{L}_{\text{wiou}} = 1 - \frac{\sum w \cdot M_{f} \cdot M_{gt} + 1}{\sum w \cdot (M_{f} + M_{gt}) - \sum w \cdot M_{f} \cdot M_{gt} + 1}.
\end{equation}
Therefore, the final loss $\mathcal{L}$ is the combination of $\mathcal{L}_{\text{wbce}}$ and $\mathcal{L}_{\text{wiou}}$ as follows:
\begin{equation}
    \mathcal{L} = \mathcal{L}_{\text{wbce}} + \lambda\mathcal{L}_{\text{wiou}},
\end{equation}
where \(\lambda\) is the weight to balance the loss terms.
\section{Experiments}
\subsection{Datasets and Evaluation Metrics}
\begin{figure}[!t]
\centering
\begin{minipage}[t]{0.48\columnwidth}
\centering
\includegraphics[width=\columnwidth]{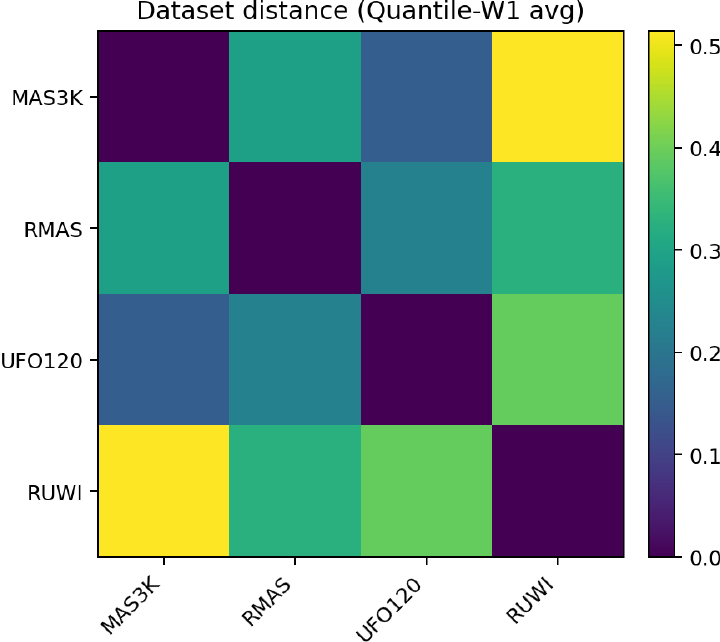}
\end{minipage}
\begin{minipage}[t]{0.48\columnwidth}
\centering
\includegraphics[width=\columnwidth]{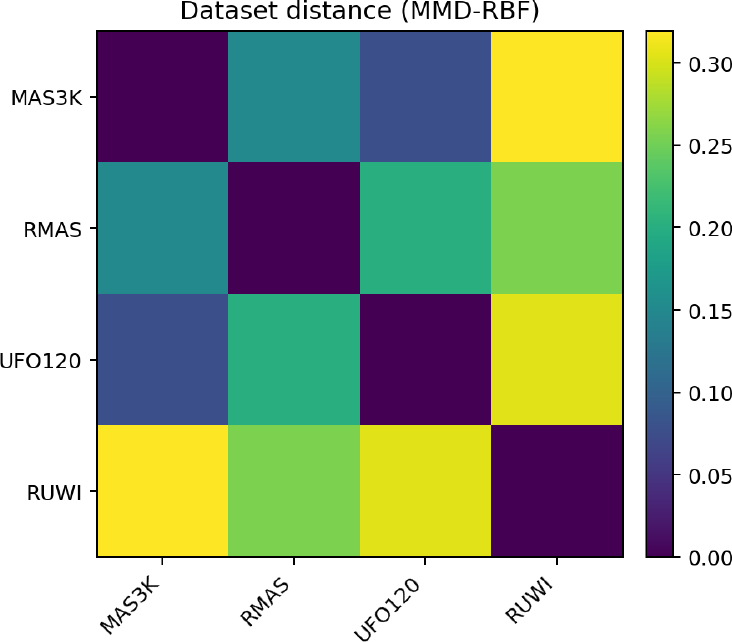}
\end{minipage}
\caption{Pairwise dataset distances measured by W1 and MMD-RBF.}
\label{fig:dataset_distance}
\end{figure}

Following previous works~\cite{zhang2024fantastic,yan2024mas}, we utilize four public MAS datasets and employ five evaluation metrics to assess the effectiveness of our model.
The \textbf{MAS3K} dataset~\cite{li2020mas3k} contains 3,103 images of marine animals, including 193 background images.
Following the standard division, we use 1,769 images for training and 1,141 images for testing.
The \textbf{RMAS} dataset~\cite{fu2023masnet} is comprised of 3,014 images of marine animals, with 2,514 images for training and 500 images for testing.
The \textbf{UFO-120} dataset~\cite{islam2020simultaneous} includes 1,620 underwater images from diverse scenes.
In line with the standard division, we employ 1,500 images for training and 120 images for testing.
The \textbf{RUWI} dataset~\cite{drews2021underwater} comprises 700 images, divided into 525 images for training and 175 images for testing.
To clarify the underwater dataset bias, we quantify pairwise dataset distances with the Wasserstein-1 (W1)~\cite{liu2025dataset} and Maximum Mean Discrepancy decoupled with Radial Basis Function (MMD-RBF)~\cite{gretton2012kernel}.
Fig.~\ref{fig:dataset_distance} shows the distance matrices.
They confirm that the bias gaps are rather large and there are non-trivial domain shifts across these datasets.
Furthermore, we conduct zero-shot experiments on a diverse set of vision tasks, including camouflaged object detection, polyp segmentation and salient object detection.
They provide additional evidence of generalization beyond these underwater datasets.

Additionally, we utilize five metrics to assess the segmentation effectiveness of various models.
They are Mean Intersection over Union (mIoU), Structural Similarity Measure ($S_\alpha$)~\cite{fan2017structure}, Weighted F-measure ($F_\beta^w$)~\cite{margolin2014evaluate}, Mean Enhanced-Alignment Measure ($mE_\phi$)~\cite{fan2021cognitive} and Mean Absolute Error (MAE).
The aforementioned five evaluation metrics allow us to comprehensively and holistically compare the strengths and weaknesses of different methods.
\subsection{Implementation Details}
We implement our model with the PyTorch toolbox and conduct experiments with one RTX 3090 GPU.
The image encoder, mask decoder and prompt encoder of SAM are initialized from the pre-trained SAM-B~\cite{kirillov2023segment}. Other parameters are randomly initialized.
During the training process, we freeze the SAM's encoder and fine-tune the remaining modules.
For fair comparison, other SAM-based methods are trained under the same protocol: we freeze the SAM's image encoder and fine-tune the remaining components/modules, using identical training/validation splits, data augmentations, and optimization settings.
The AdamW optimizer~\cite{loshchilov2017decoupled} is adopted to update the parameters.
The initial learning rate and weight decay are set to 0.001 and 0.1, respectively.
We reduce the learning rate by a factor of 10 at every 20 epochs.
Following previous methods~\cite{zhang2017amulet,fan2020pranet} the loss weight $\lambda$ is set to 1.
The total number of training epochs is set to 50.
The mini-batch size is set to 6.
All the input images are resized to $512\times512\times3$.
The reduction ratio \(r\) in FGA is set to 4.
For the evaluation, we resize the predicted masks back to the original image size by a bilinear interpolation.
The code will be released for reproduction.
\begin{figure*}[!t]
\centering
\resizebox{1.0\textwidth}{!}
{
\renewcommand\arraystretch{0.1}
\begin{tabular}{@{}c@{}c@{}c@{}c@{}c@{}c@{}c@{}c@{}c@{}c@{}c@{}c@{}c@{}c@{}c}
\vspace{0.5mm}
\includegraphics[width=0.0909\linewidth,height=1.4cm]{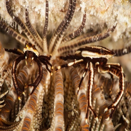}\ &
\includegraphics[width=0.0909\linewidth,height=1.4cm]{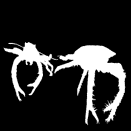}\ &
\includegraphics[width=0.0909\linewidth,height=1.4cm]{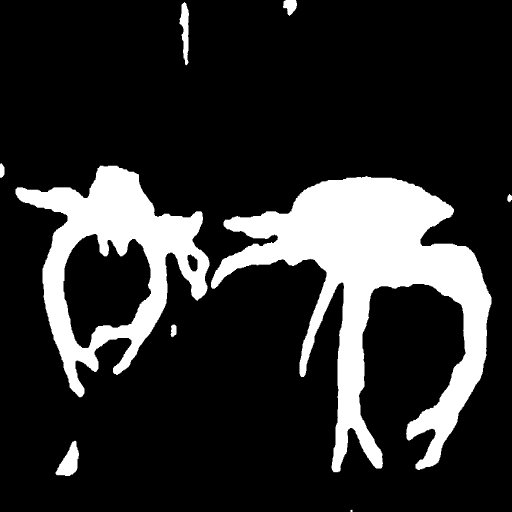}\ &
\includegraphics[width=0.0909\linewidth,height=1.4cm]{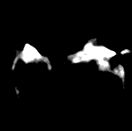}\ &
\includegraphics[width=0.0909\linewidth,height=1.4cm]{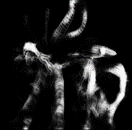}\ &
\includegraphics[width=0.0909\linewidth,height=1.4cm]{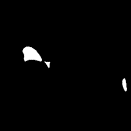}\ &
\includegraphics[width=0.0909\linewidth,height=1.4cm]{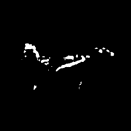}\ &
\includegraphics[width=0.0909\linewidth,height=1.4cm]{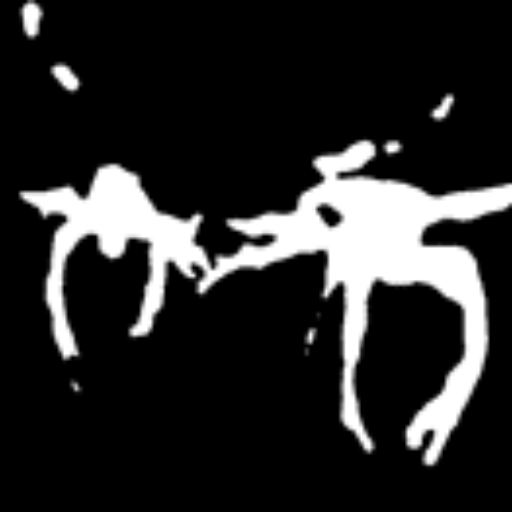}\  &
\includegraphics[width=0.0909\linewidth,height=1.4cm]{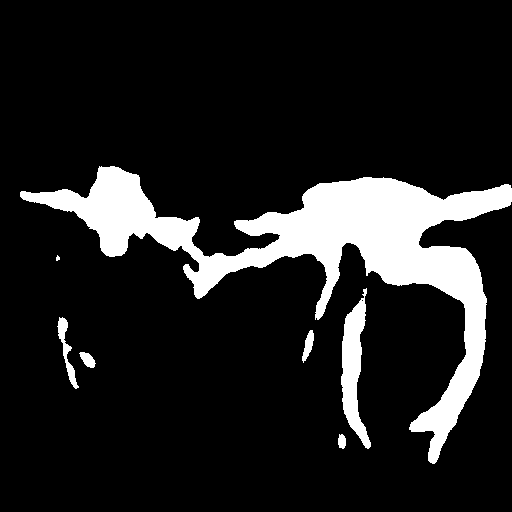}\  &
\includegraphics[width=0.0909\linewidth,height=1.4cm]{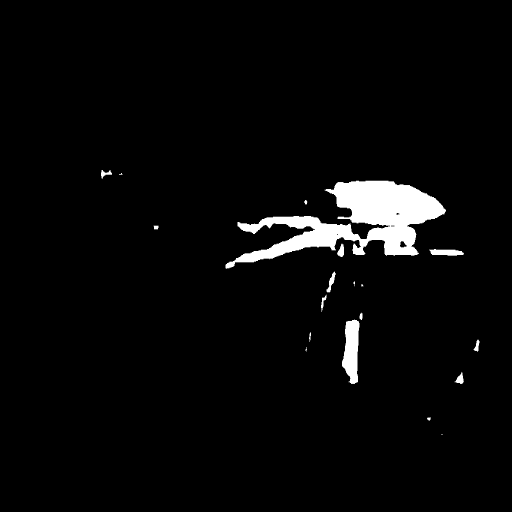}\  \\
\vspace{0.5mm}
\includegraphics[width=0.0909\linewidth,height=1.4cm]{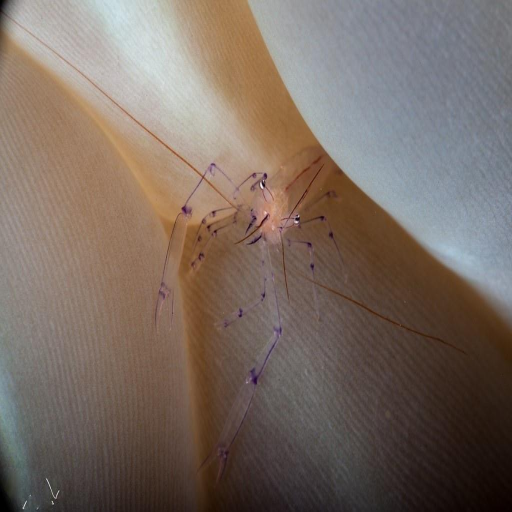}\ &
\includegraphics[width=0.0909\linewidth,height=1.4cm]{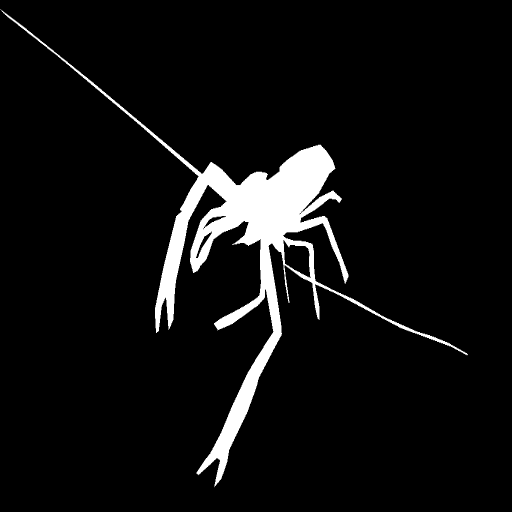}\ &
\includegraphics[width=0.0909\linewidth,height=1.4cm]{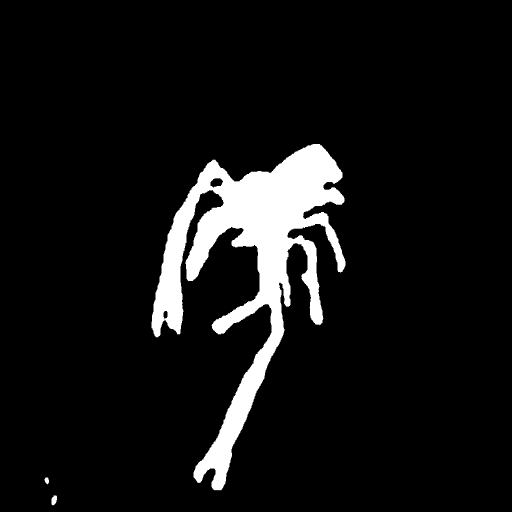}\ &
\includegraphics[width=0.0909\linewidth,height=1.4cm]{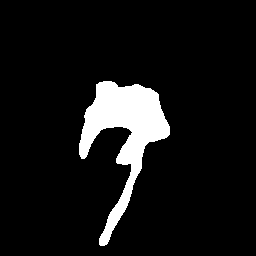}\ &
\includegraphics[width=0.0909\linewidth,height=1.4cm]{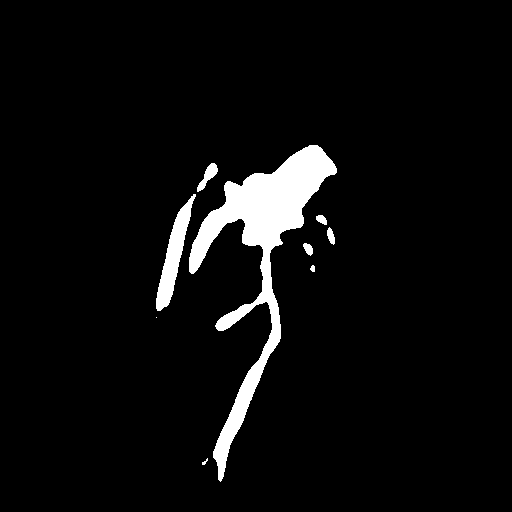}\ &
\includegraphics[width=0.0909\linewidth,height=1.4cm]{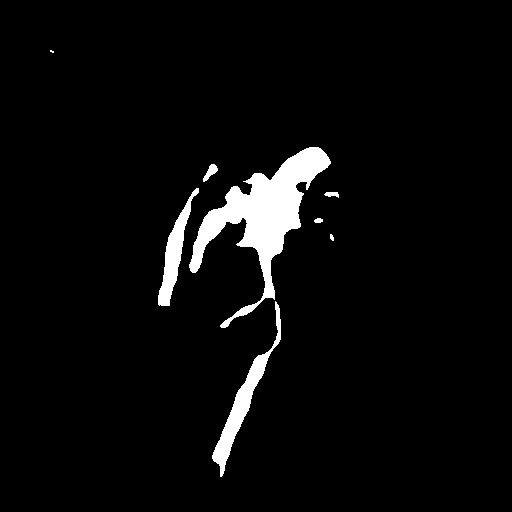}\ &
\includegraphics[width=0.0909\linewidth,height=1.4cm]{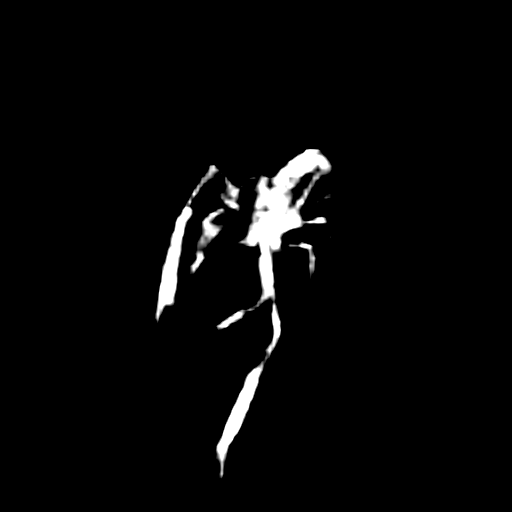}\ &
\includegraphics[width=0.0909\linewidth,height=1.4cm]{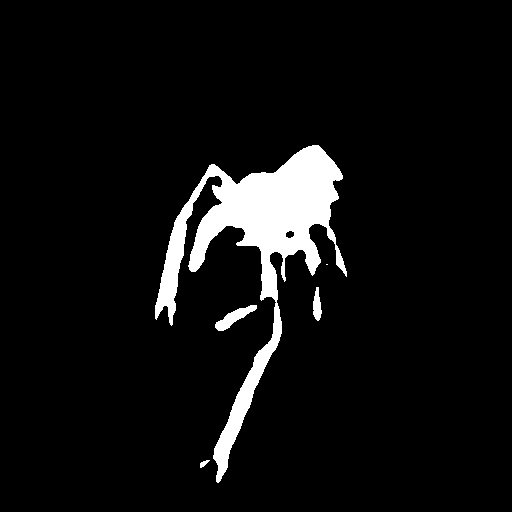}\ &
\includegraphics[width=0.0909\linewidth,height=1.4cm]{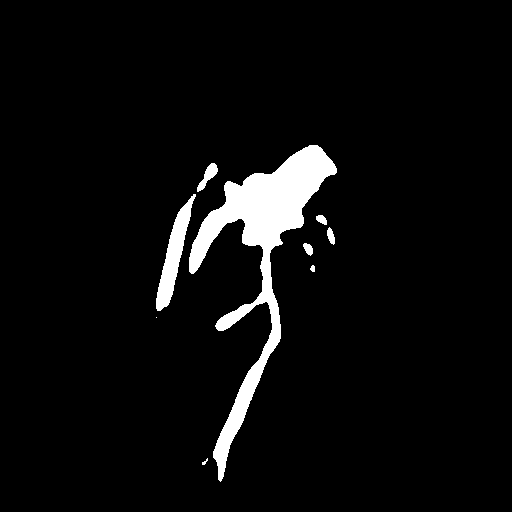}\ &
\includegraphics[width=0.0909\linewidth,height=1.4cm]{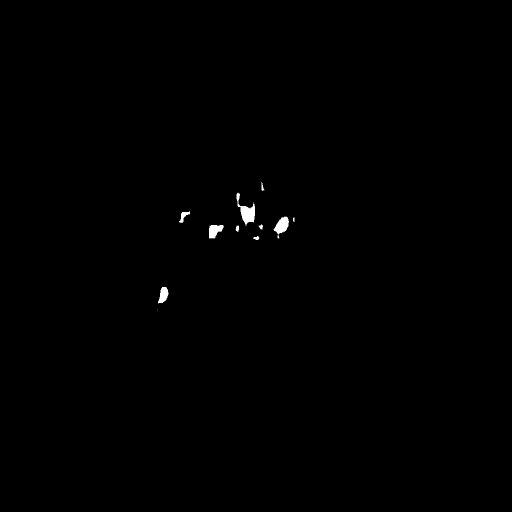}\  \\
\vspace{0.5mm}
\includegraphics[width=0.0909\linewidth,height=1.4cm]{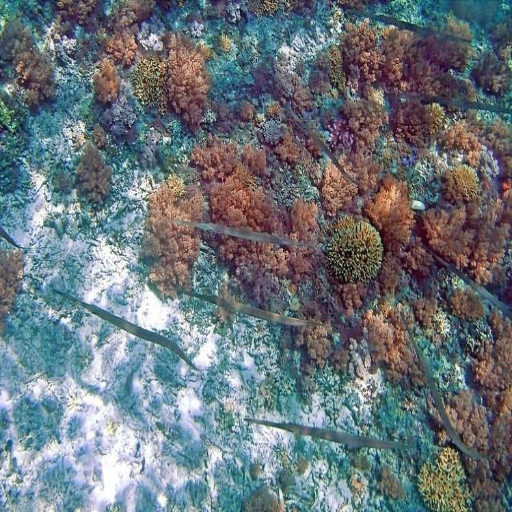}\ &
\includegraphics[width=0.0909\linewidth,height=1.4cm]{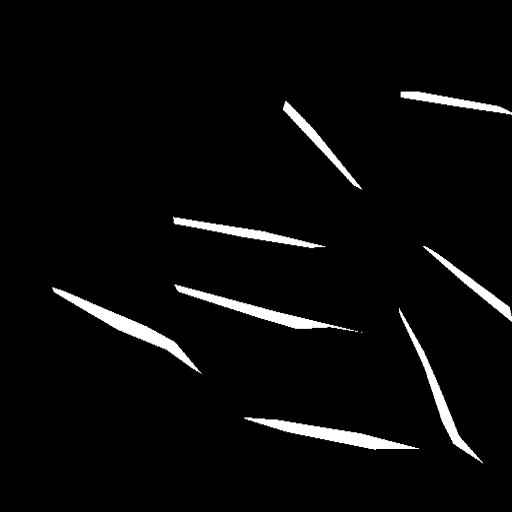}\ &
\includegraphics[width=0.0909\linewidth,height=1.4cm]{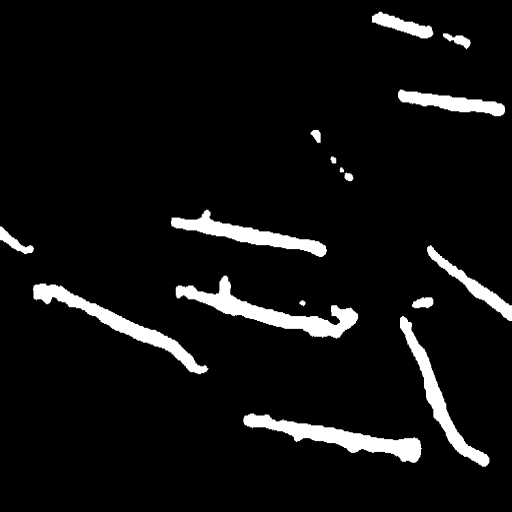}\ &
\includegraphics[width=0.0909\linewidth,height=1.4cm]{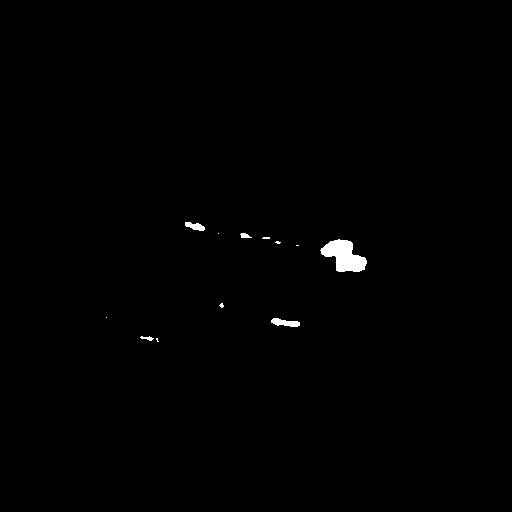}\ &
\includegraphics[width=0.0909\linewidth,height=1.4cm]{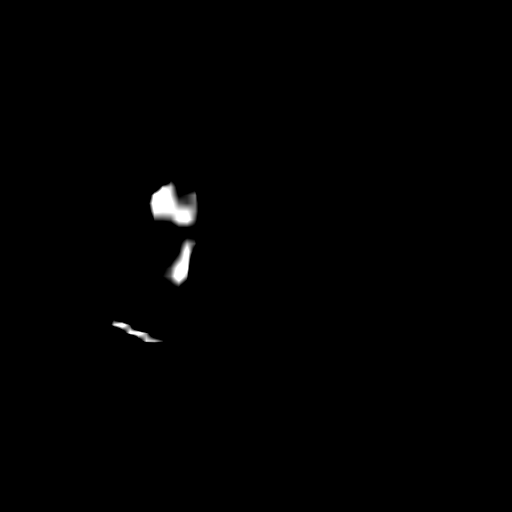}\ &
\includegraphics[width=0.0909\linewidth,height=1.4cm]{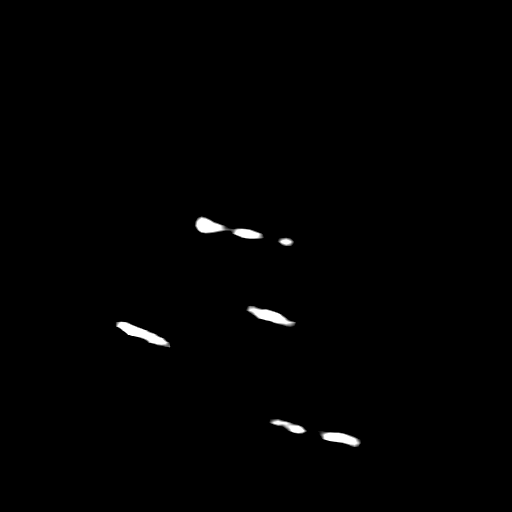}\ &
\includegraphics[width=0.0909\linewidth,height=1.4cm]{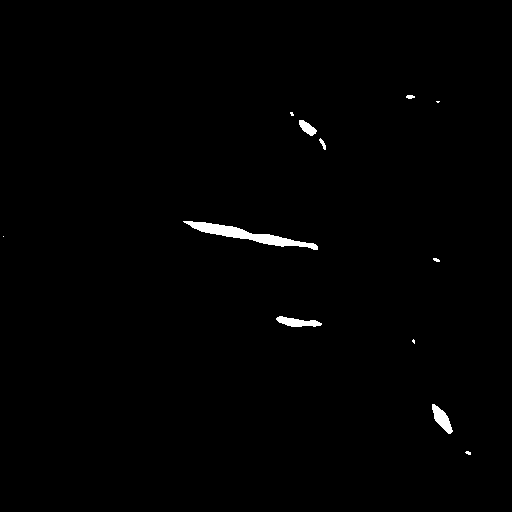}\ &
\includegraphics[width=0.0909\linewidth,height=1.4cm]{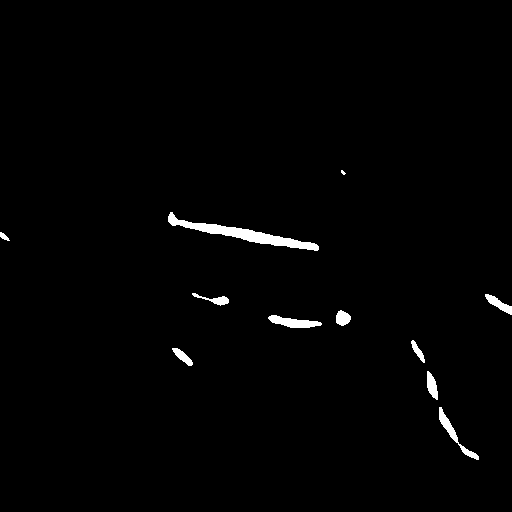}\ &
\includegraphics[width=0.0909\linewidth,height=1.4cm]{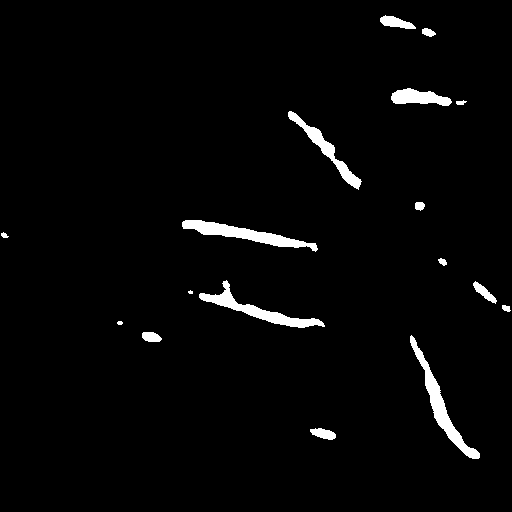}\ &
\includegraphics[width=0.0909\linewidth,height=1.4cm]{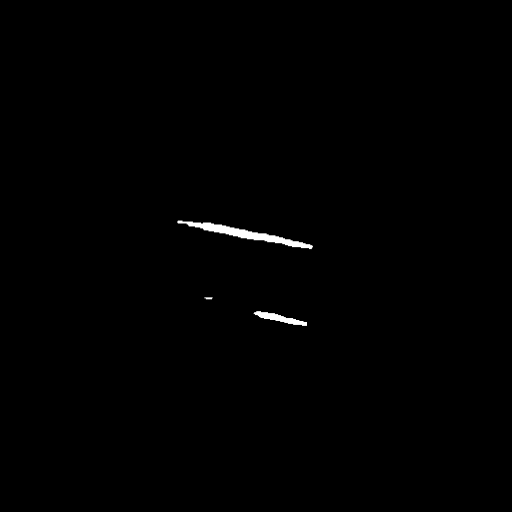}\ &
\\
\vspace{0.5mm}
\includegraphics[width=0.0909\linewidth,height=1.4cm]{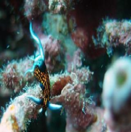}\ &
\includegraphics[width=0.0909\linewidth,height=1.4cm]{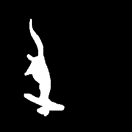}\ &
\includegraphics[width=0.0909\linewidth,height=1.4cm]{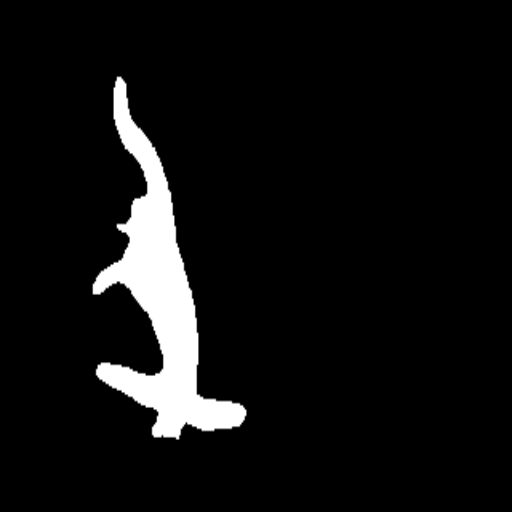}\ &
\includegraphics[width=0.0909\linewidth,height=1.4cm]{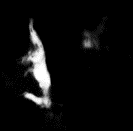}\ &
\includegraphics[width=0.0909\linewidth,height=1.4cm]{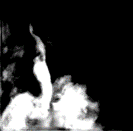}\ &
\includegraphics[width=0.0909\linewidth,height=1.4cm]{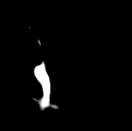}\ &
\includegraphics[width=0.0909\linewidth,height=1.4cm]{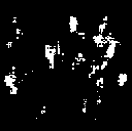}\ &
\includegraphics[width=0.0909\linewidth,height=1.4cm]{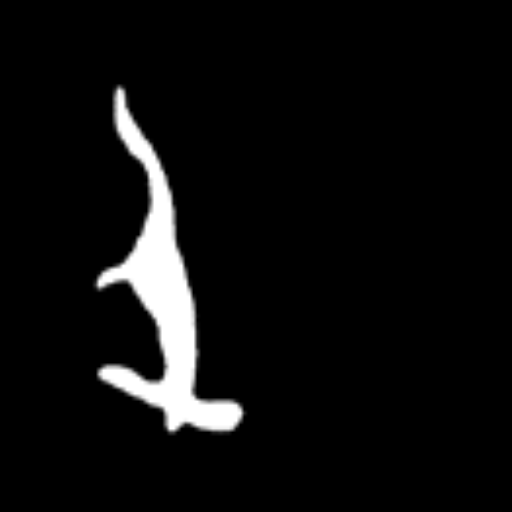}\ &
\includegraphics[width=0.0909\linewidth,height=1.4cm]{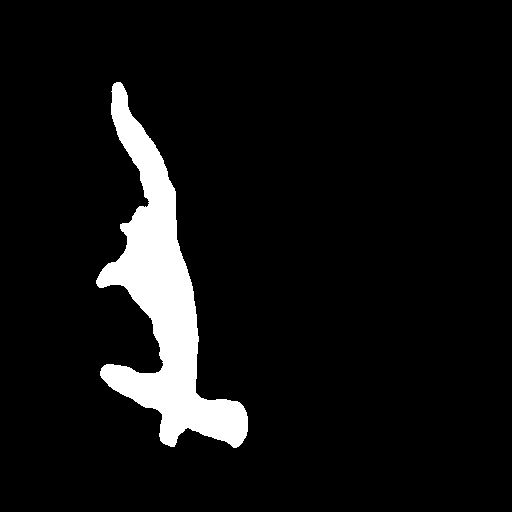}\ &
\includegraphics[width=0.0909\linewidth,height=1.4cm]{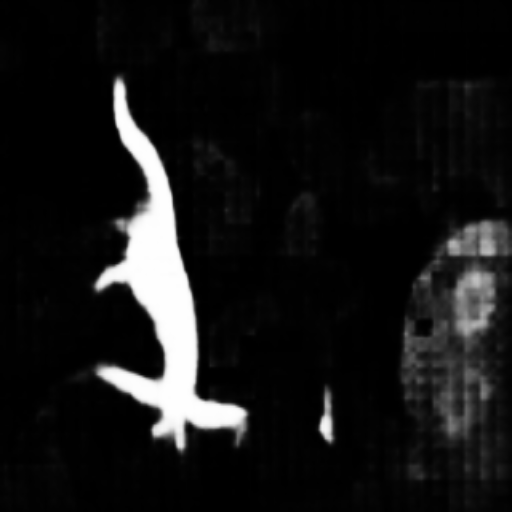}\ &
\\
\vspace{0.5mm}
\includegraphics[width=0.0909\linewidth,height=1.4cm]{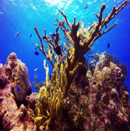}\ &
\includegraphics[width=0.0909\linewidth,height=1.4cm]{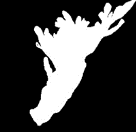}\ &
\includegraphics[width=0.0909\linewidth,height=1.4cm]{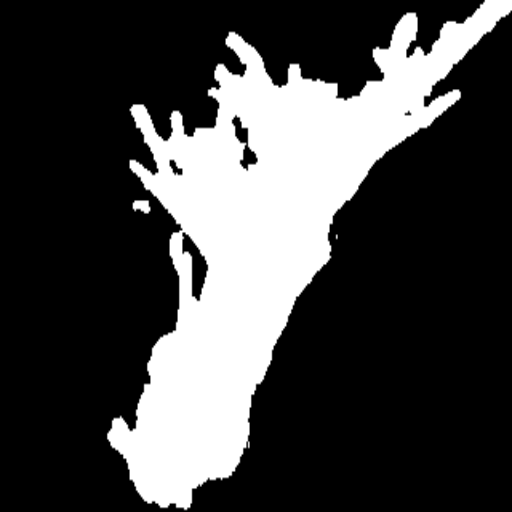}\ &
\includegraphics[width=0.0909\linewidth,height=1.4cm]{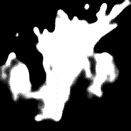}\ &
\includegraphics[width=0.0909\linewidth,height=1.4cm]{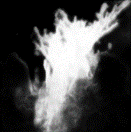}\ &
\includegraphics[width=0.0909\linewidth,height=1.4cm]{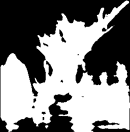}\ &
\includegraphics[width=0.0909\linewidth,height=1.4cm]{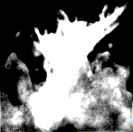}\ &
\includegraphics[width=0.0909\linewidth,height=1.4cm]{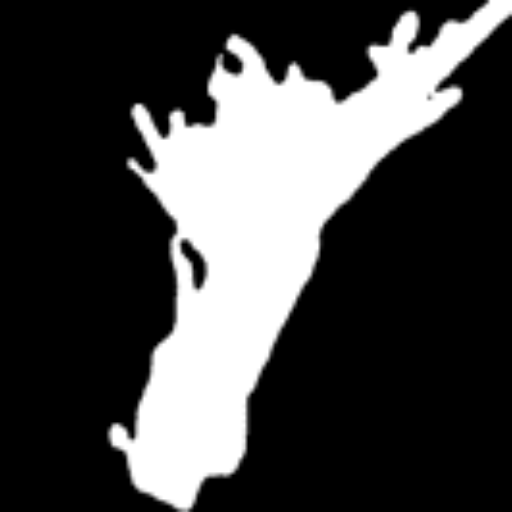}\ &
\includegraphics[width=0.0909\linewidth,height=1.4cm]{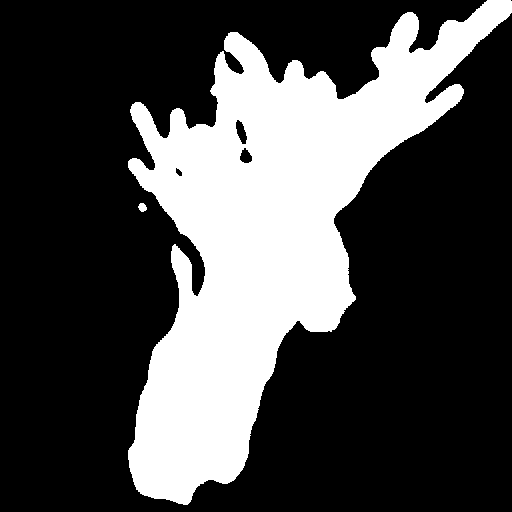}\ &
\includegraphics[width=0.0909\linewidth,height=1.4cm]{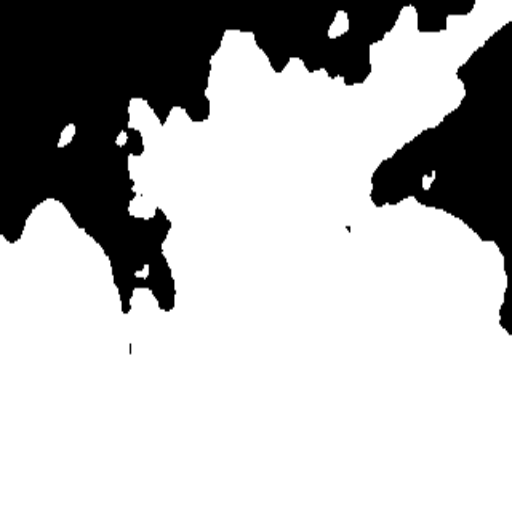}\ &
\\
\vspace{0.5mm}
 (a) & (b) & (c)  & (d) & (e) & (f) & (g) & (h) & (i) & (j) \\
\end{tabular}
}
\caption{Visual comparison of predicted segmentation masks with different methods. From left to right: (a) \textbf{Input Images}; (b) \textbf{Ground Truth}; (c) \textbf{Ours}; (d) \textbf{MASNet}; (e) \textbf{SETR}; (f) \textbf{TransUNet}; (g) \textbf{SAM}; (h) \textbf{MAS-SAM}; (i) \textbf{Dual-SAM}; (j) \textbf{SAM2}.}
\label{fig:visual}
\end{figure*}
\begin{figure*}[!t]
\centering
\resizebox{1.0\textwidth}{!}
{
\renewcommand\arraystretch{0.1}
\begin{tabular}{@{}c@{}c@{}c@{}c@{}c@{}c@{}c@{}c@{}c@{}c@{}c@{}c@{}c@{}c}
\vspace{0.5mm}
\includegraphics[width=0.125\linewidth,height=1.4cm]{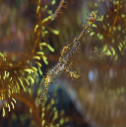}\ &
\includegraphics[width=0.125\linewidth,height=1.4cm]{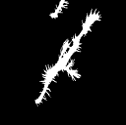}\ &
\includegraphics[width=0.125\linewidth,height=1.4cm]{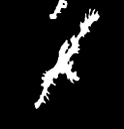}\ &
\includegraphics[width=0.125\linewidth,height=1.4cm]{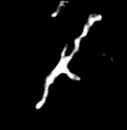}\ &
\includegraphics[width=0.125\linewidth,height=1.4cm]{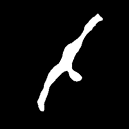}\ &
\includegraphics[width=0.125\linewidth,height=1.4cm]{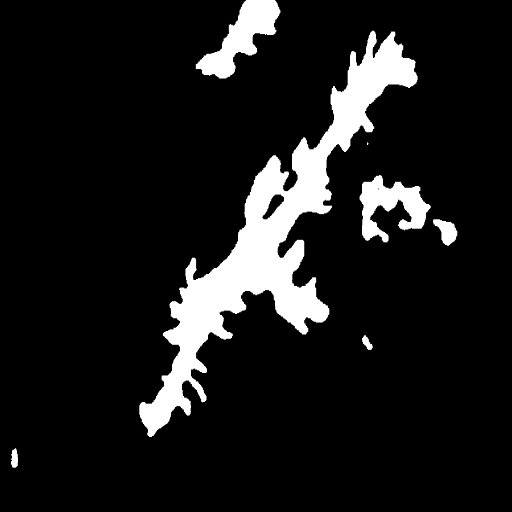}\ &
\includegraphics[width=0.125\linewidth,height=1.4cm]{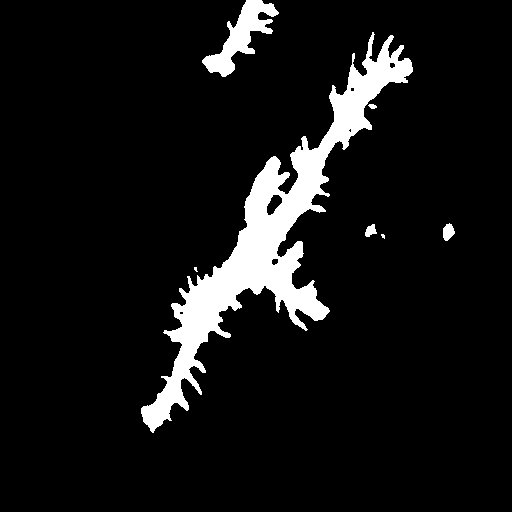}\ &
\includegraphics[width=0.125\linewidth,height=1.4cm]{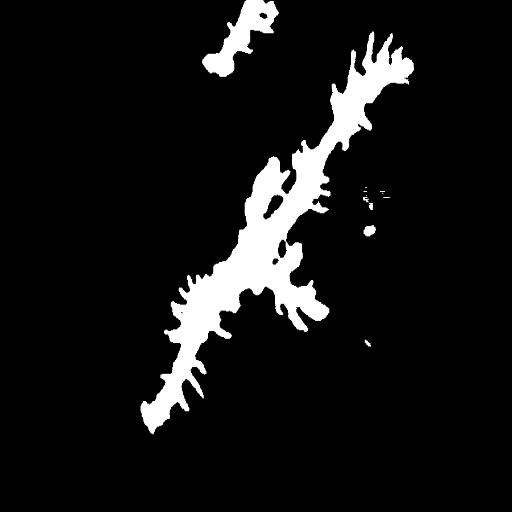}\ &
\includegraphics[width=0.125\linewidth,height=1.4cm]{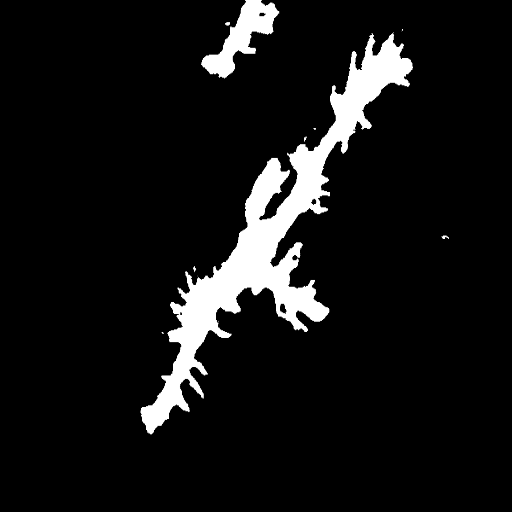}\  \\
\vspace{0.5mm}
\includegraphics[width=0.125\linewidth,height=1.4cm]{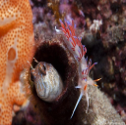}\ &
\includegraphics[width=0.125\linewidth,height=1.4cm]{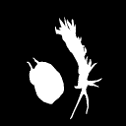}\ &
\includegraphics[width=0.125\linewidth,height=1.4cm]{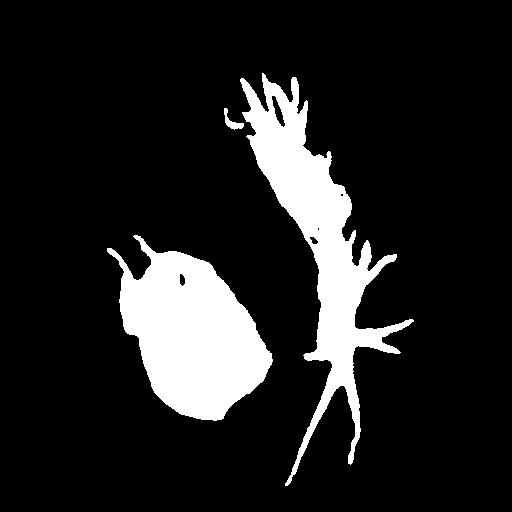}\ &
\includegraphics[width=0.125\linewidth,height=1.4cm]{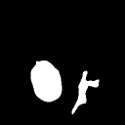}\ &
\includegraphics[width=0.125\linewidth,height=1.4cm]{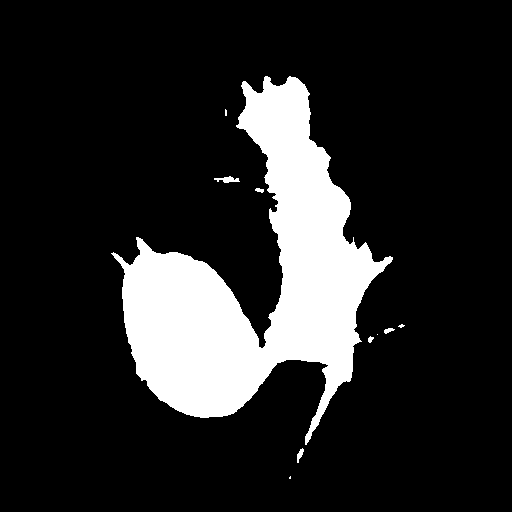}\ &
\includegraphics[width=0.125\linewidth,height=1.4cm]{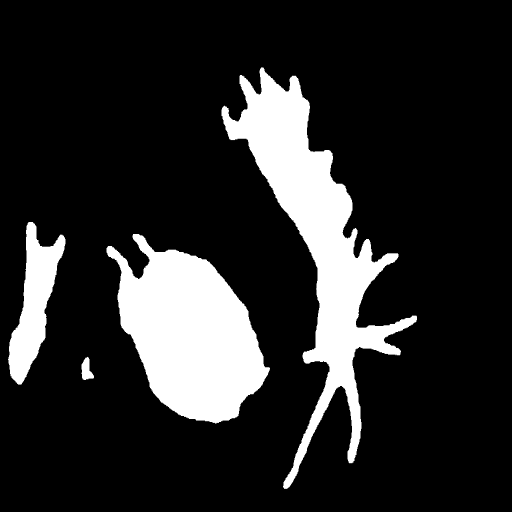}\ &
\includegraphics[width=0.125\linewidth,height=1.4cm]{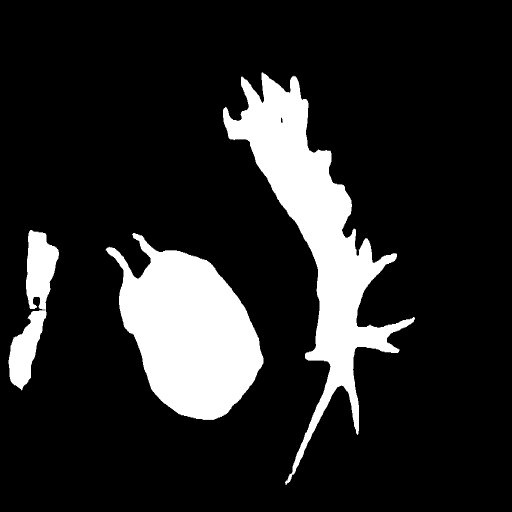}\ &
\includegraphics[width=0.125\linewidth,height=1.4cm]{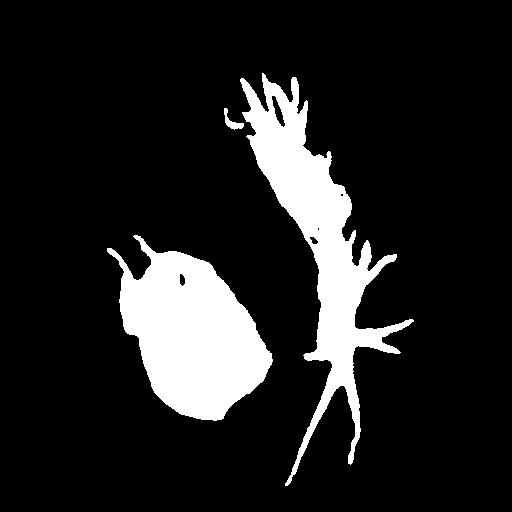}\ &
\includegraphics[width=0.125\linewidth,height=1.4cm]{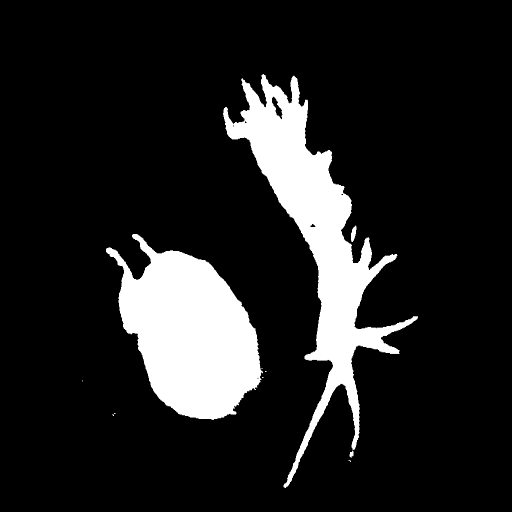}\  \\
 Image &  GT &  HFP-SAM& (A) &  (B)  &  (C) &  (D) &  (E)&(F)  \\
\end{tabular}
}
\caption{Visual comparison of predicted segmentation masks with different modules from Tab.~\ref{ablation}.}
\label{fig:ablation}
\end{figure*}
\subsection{Comparison with the State-of-the-arts}
We compare our method with other state-of-the-art methods.
Firstly, we compare with the latest CNN-based MAS methods~\cite{fu2023masnet,li2021marine}.
We follow the compared methods and metrics provided by MASNet~\cite{fu2023masnet}.
Additionally, we provide representative Transformer-based methods~\cite{zheng2021rethinking,chen2021transunet,he2023h2former}.
Furthermore, we compare with other advanced SAM-based techniques~\cite{yan2024mas,zhang2024fantastic,lai2023detect,chen2024sam2,xiong2024sam2,ravi2024sam} that customize SAM for downstream tasks.
Both quantitative and qualitative results demonstrate the superiority of our proposed method.

\textbf{Quantitative Comparisons.}
Tab.~\ref{mas3k} and Tab.~\ref{ufo} present the quantitative results of all compared methods.
Our method surpasses others across all five metrics on four datasets.
This evidence highlights our method's exceptional performance.

When compared with CNN-based methods, our method exhibits much better.
On the large-scale MAS3K dataset, our method achieves the best mIoU, $S_\alpha$ , $F_\beta^w$ and $mE_\phi$ values.
This demonstrates an enhancement of roughly 8-10\% across diverse metrics.
Similarly, our method shows comparable performance on the other three datasets.
This indicates that our method indeed surpasses CNN-based methods in acquiring semantic information via a global perspective.

When compared with Transformer-based methods, our method achieves an increase of 6-8\% in metrics on the MAS3K dataset.
Furthermore, notable improvements are achieved across other datasets as well.
Consequently, our method delivers considerable benefits over other methods, leveraging SAM's image segmentation capabilities.

When compared with most SAM-based methods, our method achieves a 4-6\% performance improvement.
By utilizing adapters and prompts informed by frequency domain priors, our method effectively infuses domain-specific information into SAM.
Additionally, the deployment of FVM in SAM's decoder captures fine-grained details.
This module significantly boosts the SAM's segmentation capabilities.

\textbf{Qualitative Comparisons.}
To vividly show the advantages of our method, we offer visual comparisons across various methods in Fig.~\ref{fig:visual}.
These visual examples clearly illustrate our method's superiority over previous techniques.
The images pose significant challenges due to their highly cluttered backgrounds and intricate details.
Our method consistently produces superior segmentation masks, highlighting its robust performance in complex scenarios.
\subsection{Ablation Study}
Ablation experiments are performed to assess the impact of key components within our framework.
These evaluations are carried out on the MAS3K dataset.
Regarding model (A), it leverages parameters from pre-training SAM-B~\cite{kirillov2023segment} and subsequently fine-tunes the decoder to enhance its specificity for the MAS task.
The difference between (B) and (C) lies in whether a frequency domain prior mask is used to guide the standard adapter.
More detailed visual results and comparisons can be found in Fig.~\ref{fig:ablation}.
\begin{table}[h]
\centering
\caption{Performance with different key modules on MAS3K.}
\resizebox{0.5\textwidth}{!}
{
\begin{tabular}{c|ccccc|c|c|c|c|c}
        \hline
        &\multicolumn{5}{c|}{\textbf{Module}}        &\multicolumn{5}{c}{\textbf{MAS3K}} \\ \cline{2-11}
        & \textbf{Adapter}& \textbf{FGA}& \textbf{FPS}& \textbf{FVM}&\textbf{SL} & \textbf{mIoU} & $\textbf{S}_\alpha$ & $\textbf{F}_\beta^w$ & $\textbf{m}\textbf{E}_\phi$ & \textbf{MAE} \\
        \hline
        (A) & \ding{53}& \ding{53}&\ding{53}& \ding{53}& \ding{53}& 0.566 & 0.763 & 0.656 & 0.807& 0.059 \\
        (B) & \ding{51}& \ding{53}&\ding{53}& \ding{53}& \ding{53}& 0.739 & 0.860 & 0.810& 0.917& 0.031 \\
        (C) & \ding{51}& \ding{51}&\ding{53}& \ding{53}& \ding{53}& 0.754 & 0.865 & 0.818 & 0.923& 0.030 \\
        (D) & \ding{51}& \ding{51}& \ding{51}&\ding{53}& \ding{53}& 0.771 & 0.872 & 0.827 & 0.929& 0.028\\
        (E) & \ding{51}& \ding{51}&\ding{51}& \ding{51}& \ding{53}& 0.792&0.881&	0.839&	0.932& 0.026 \\
        (F) & \ding{51}& \ding{51}&\ding{51}& \ding{51}& \ding{51}& 0.797 & 0.888 & 0.845 & 0.938& 0.024 \\
        \hline
\end{tabular}
}
\label{ablation}
\end{table}
\begin{figure}[!h]
\centering
\resizebox{0.46\textwidth}{!}
{
\renewcommand\arraystretch{0.1}
\begin{tabular}{@{}c@{}c@{}c}
\vspace{0.5mm}
\includegraphics[width=0.33\linewidth,height=1.6cm]{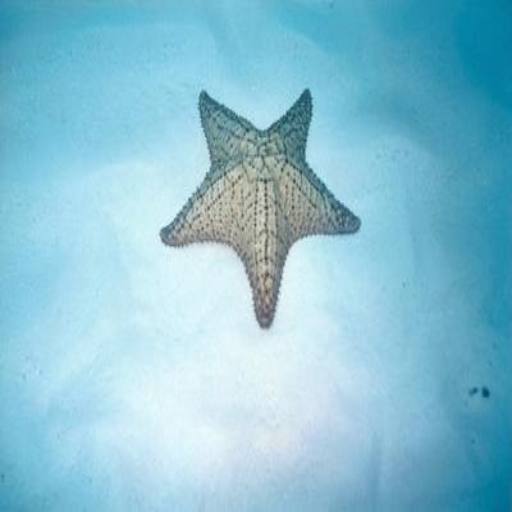}\ &
\includegraphics[width=0.33\linewidth,height=1.6cm]{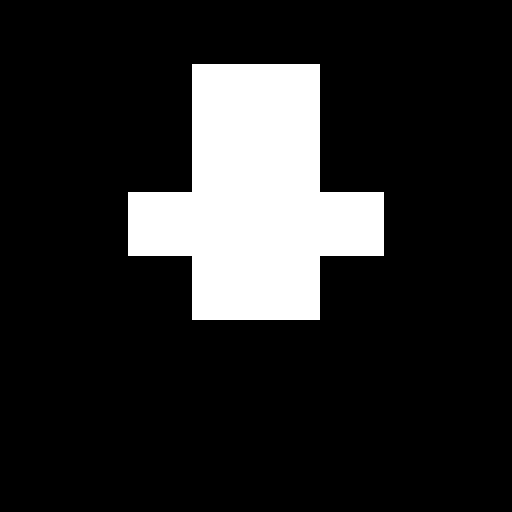}\ &
\includegraphics[width=0.33\linewidth,height=1.6cm]{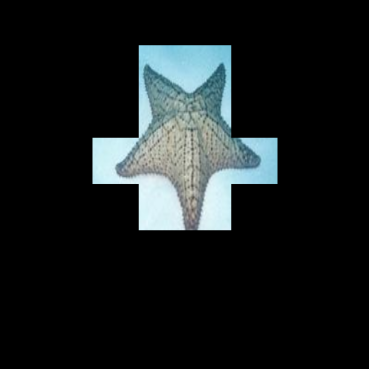}\ \\
\vspace{0.5mm}
\includegraphics[width=0.33\linewidth,height=1.6cm]{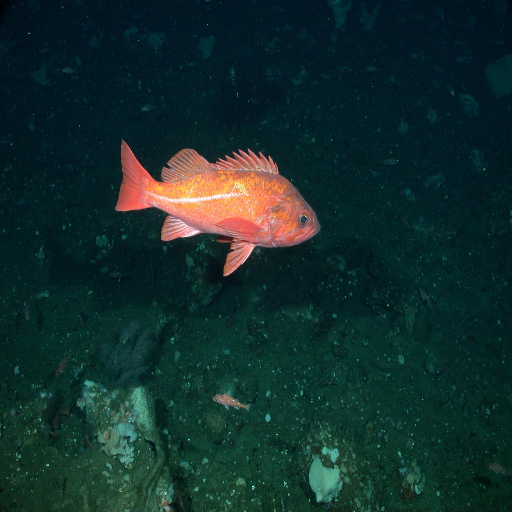}\ &
\includegraphics[width=0.33\linewidth,height=1.6cm]{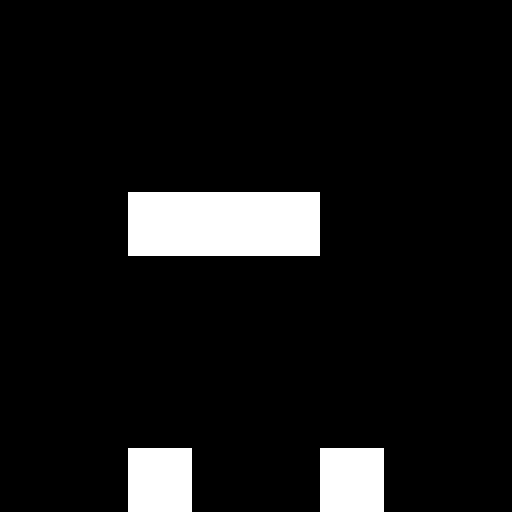}\ &
\includegraphics[width=0.33\linewidth,height=1.6cm]{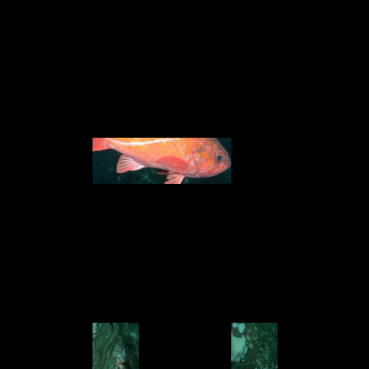}\ \\
\vspace{0.5mm}
 (a)  & (b) &(c)  \\
\end{tabular}
}
\caption{Visual effect of using FGA. (a) Input Images; (b) Windows selected by FGA; (c) Images after applying the window masks.}
\label{fga_explain}
\end{figure}
\begin{figure}[!h]
\centering
\resizebox{0.46\textwidth}{!}
{
\renewcommand\arraystretch{0.1}
\begin{tabular}{@{}c@{}c@{}c@{}c}
\vspace{0.5mm}
\includegraphics[width=0.25\linewidth,height=1.35cm]{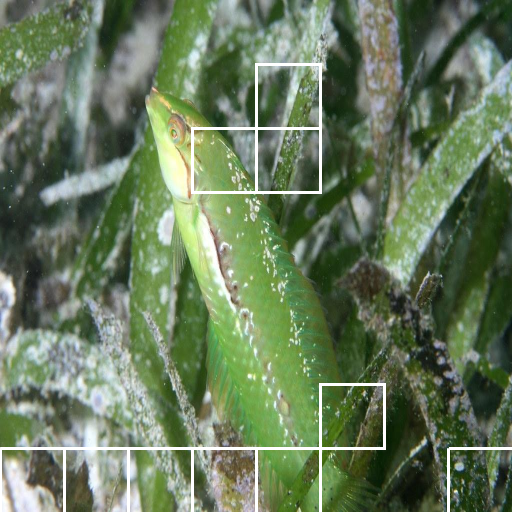}\ &
\includegraphics[width=0.25\linewidth,height=1.35cm]{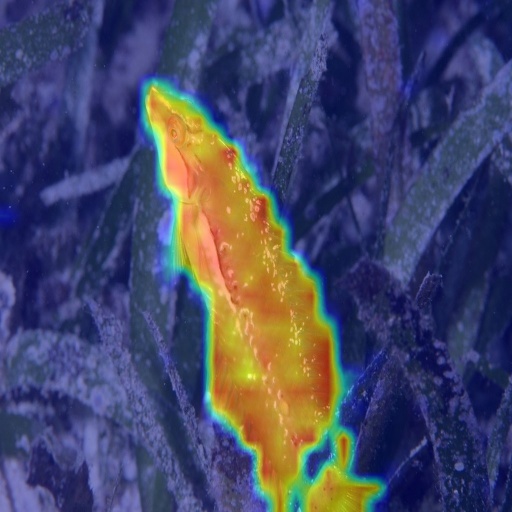}\ &
\includegraphics[width=0.25\linewidth,height=1.35cm]{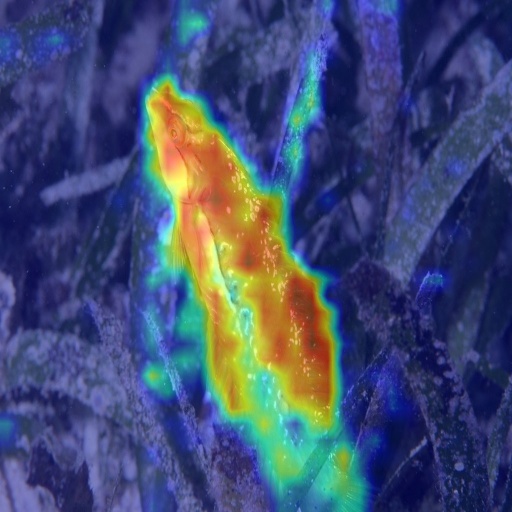}\ &
\includegraphics[width=0.25\linewidth,height=1.35cm]{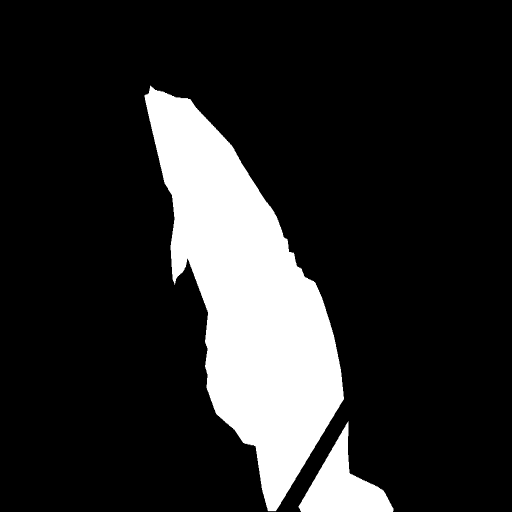}\ \\
\vspace{0.5mm}
\includegraphics[width=0.25\linewidth,height=1.35cm]{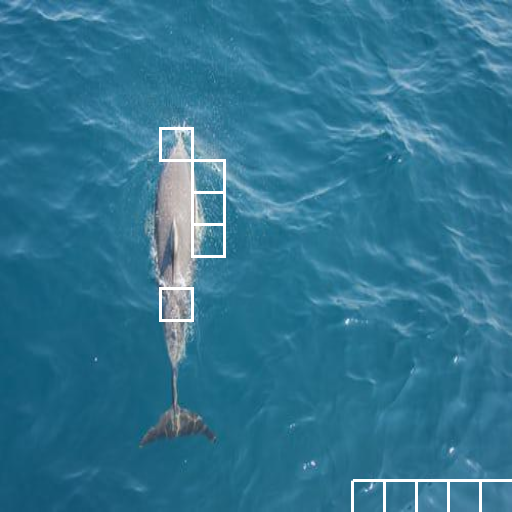}\ &
\includegraphics[width=0.25\linewidth,height=1.35cm]{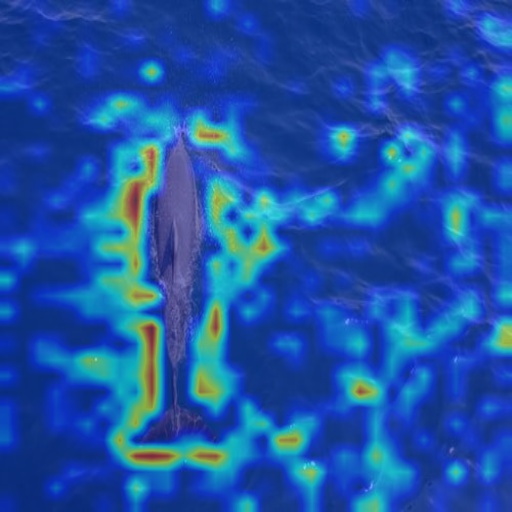}\ &
\includegraphics[width=0.25\linewidth,height=1.35cm]{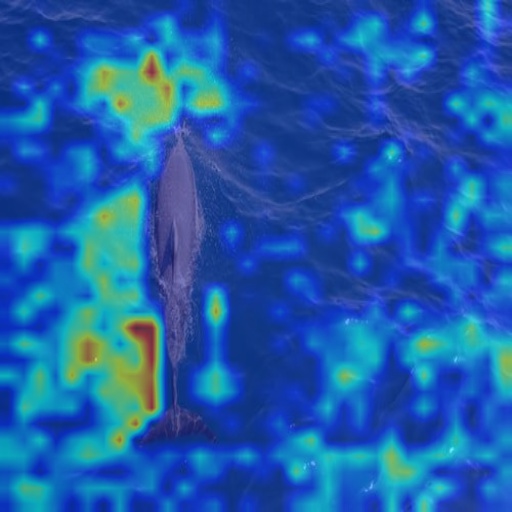}\ &
\includegraphics[width=0.25\linewidth,height=1.35cm]{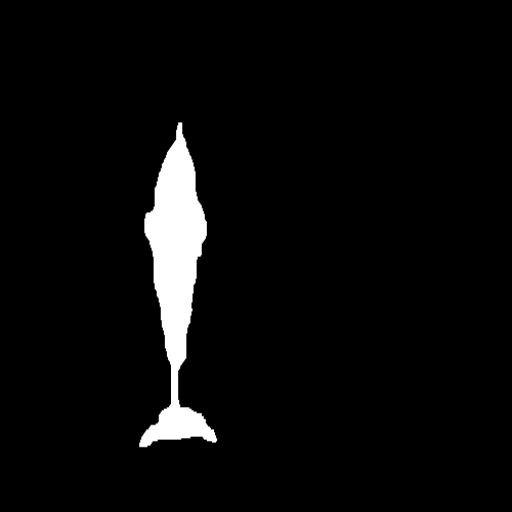}\ \\
\vspace{0.5mm}
 (a) &  (b) &(c) & (d) \\
\end{tabular}
}
\caption{(a) shows the regions with significant frequency domain changes captured by the FGA. (b) and (c) are the feature maps with the FGA and a standard adapter. (d) is the ground truth. Best view by zooming in.}
\label{fga_visual}
\end{figure}
\begin{figure*}[!t]
\centering
\resizebox{1.0\textwidth}{!}
{
\renewcommand\arraystretch{0.1}
\begin{tabular}{@{}c@{}c@{}c@{}c@{}c@{}c@{}c@{}c@{}c@{}c@{}c@{}c@{}c}
\vspace{0.5mm}
\includegraphics[width=0.125\linewidth,height=1.4cm]{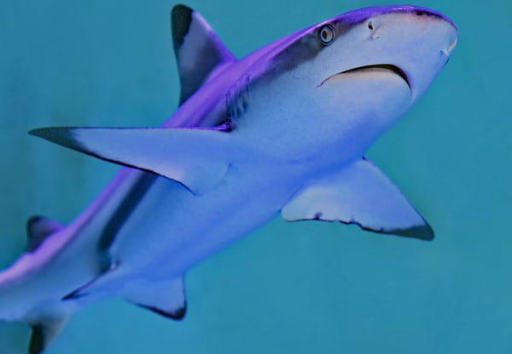}\ &
\includegraphics[width=0.125\linewidth,height=1.4cm]{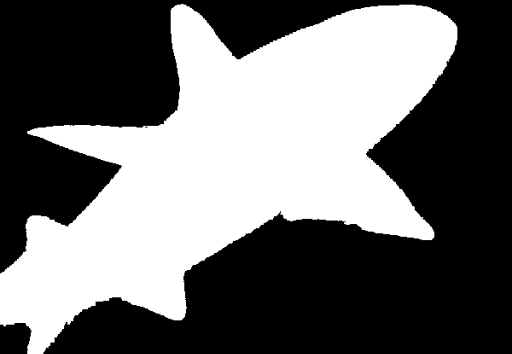}\ &
\includegraphics[width=0.125\linewidth,height=1.4cm]{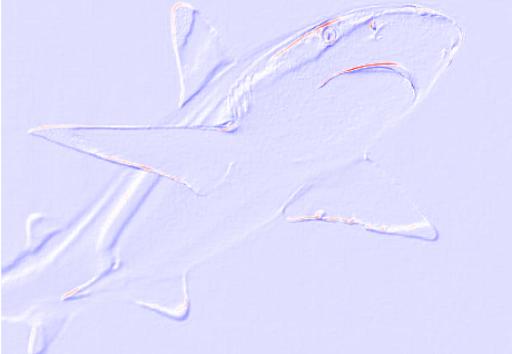}\ &
\includegraphics[width=0.125\linewidth,height=1.4cm]{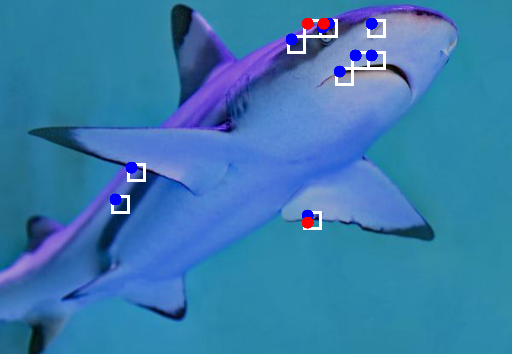}\ &
\includegraphics[width=0.125\linewidth,height=1.4cm]{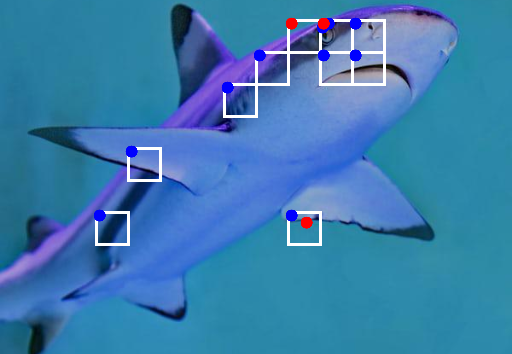}\ &
\includegraphics[width=0.125\linewidth,height=1.4cm]{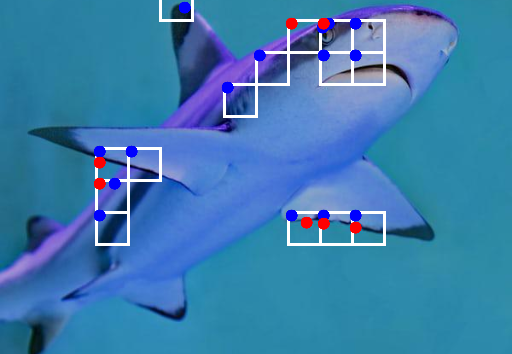}\ &
\includegraphics[width=0.125\linewidth,height=1.4cm]{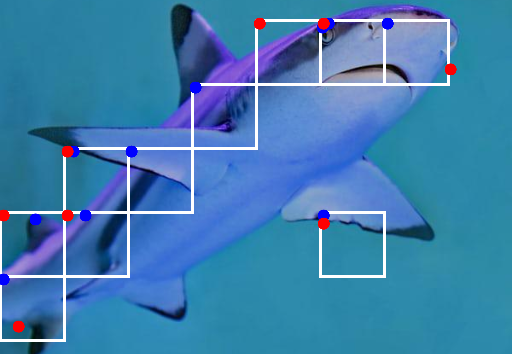}\ &\\
\vspace{0.5mm}
\includegraphics[width=0.125\linewidth,height=1.4cm]{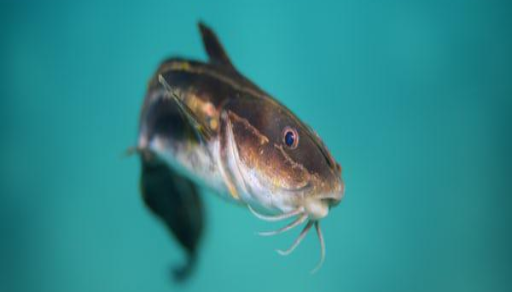}\ &
\includegraphics[width=0.125\linewidth,height=1.4cm]{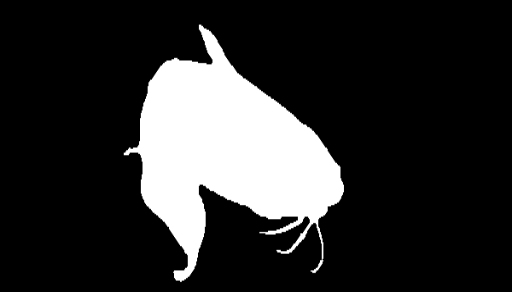}\ &
\includegraphics[width=0.125\linewidth,height=1.4cm]{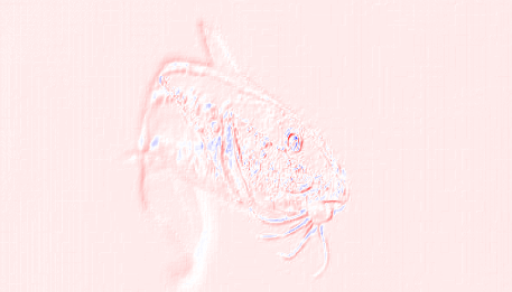}\ &
\includegraphics[width=0.125\linewidth,height=1.4cm]{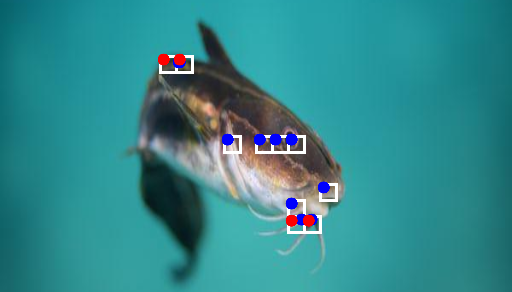}\ &
\includegraphics[width=0.125\linewidth,height=1.4cm]{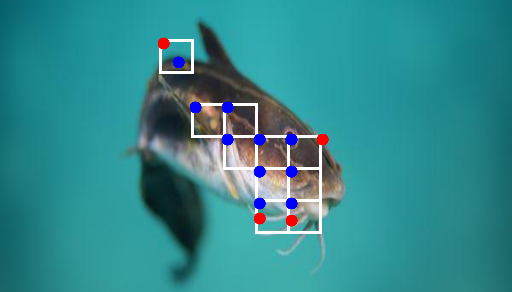}\ &
\includegraphics[width=0.125\linewidth,height=1.4cm]{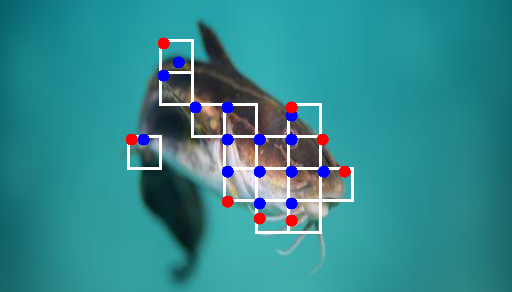}\ &
\includegraphics[width=0.125\linewidth,height=1.4cm]{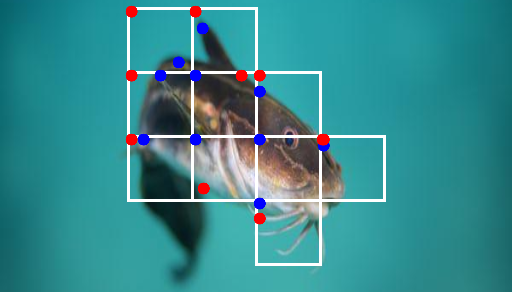}\ &\\
\vspace{0.5mm}
Image &  GT &   Frequency& (a)&(b)&(c)&(d)\\
\end{tabular}
}
\caption{Visual comparison of different amounts and sizes of sliding windows in FPS. The ablation settings from left to right: (a) \textbf{Window=10, Size=16}; (b) \textbf{Window=10, Size=32}; (c) \textbf{Window=15, Size=32}; (d) \textbf{Window=10, Size=64}.}
\label{fig:fre_visual}
\end{figure*}

\textbf{Effects of FGA.}
In rows 2-3 and 5 of Tab.~\ref{ablation}, we demonstrate the effectiveness of FGA.
Our FGA results in an improvement of approximately 1\% in mIoU.
Fig.~\ref{fga_explain} shows the windows selected by FGA.
Fig.~\ref{fga_visual} visually demonstrates the benefits of using FGA.
It is clear that our FGA focuses on key areas by utilizing frequency domain information.
\begin{table}[h!]
\centering
\caption{Performance with different settings in FPS.}
\resizebox{0.46\textwidth}{!}
{
\begin{tabular}{ccc|c|c|c|c}
\hline
\multicolumn{1}{c}{Window}               & \multicolumn{1}{c}{Size}                & Point             & \textbf{mIoU}  & $\textbf{F}^w_{\beta}$ & $\textbf{mE}_{\phi}$ &\textbf{Time}(s)  \\ \hline
\multicolumn{1}{c}{5}                   & \multicolumn{1}{c}{\multirow{3}{*}{32}} & \multirow{3}{*}{2} & 0.767        & 0.825    & 0.927     &0.0087  \\
\multicolumn{1}{c}{10}                  & \multicolumn{1}{c}{}                    &                    & 0.771  & 0.827 & 0.929  &0.0093 \\
\multicolumn{1}{c}{20}                  & \multicolumn{1}{c}{}                    &                    & 0.772       & 0.827    & 0.930      &0.0117 \\ \hline
\multicolumn{1}{c}{\multirow{3}{*}{10}} & \multicolumn{1}{c}{16}                  & \multirow{3}{*}{2} & 0.762        & 0.823    & 0.926    &0.0090   \\
\multicolumn{1}{c}{}                    & \multicolumn{1}{c}{32}                  &                    & 0.771  & 0.827 & 0.929&0.0093 \\
\multicolumn{1}{c}{}                    & \multicolumn{1}{c}{64}                  &                    & 0.766       & 0.824    & 0.927     &0.0106 \\ \hline
\multicolumn{1}{c}{\multirow{3}{*}{10}} & \multicolumn{1}{c}{\multirow{3}{*}{32}} & 1                  & 0.768        & 0.825    & 0.928    &0.0089   \\
\multicolumn{1}{c}{}                    & \multicolumn{1}{c}{}                    & 2                  & 0.771  & 0.827 & 0.929& 0.0093  \\
\multicolumn{1}{c}{}                    & \multicolumn{1}{c}{}                    & 4                  & 0.771        & 0.828    & 0.929   &0.0112  \\ \hline

\end{tabular}
}
\label{FPS}
\end{table}
\begin{table}[h!]
\centering
\caption{Performance with different point prompt methods.}
\resizebox{0.44\textwidth}{!}
{
\begin{tabular}{ccc|c|c|c|c}
\hline
\multicolumn{1}{c}{}& \multicolumn{1}{c}{\textbf{Method}}&& \textbf{mIoU}&$\textbf{F}^w_{\beta}$ & $\textbf{mE}_{\phi}$ &\textbf{Time}(s)  \\
\hline
\multicolumn{3}{c|}{Random Sampling} & 0.760& 0.823& 0.925& 0.0013\\
\multicolumn{3}{c|}{Global Sampling} & 0.764& 0.824& 0.926& 0.3148\\
\multicolumn{3}{c|}{FPS}             & 0.771& 0.827& 0.929& 0.0093\\
\hline
\end{tabular}
}
\label{select_mode}
\end{table}
\begin{table}[h!]
\centering
\caption{Performance with different point prompts in FPS.}
\resizebox{0.44\textwidth}{!}
{
\begin{tabular}{l|c|c|c|c|c}
\hline
\textbf{Method} & \textbf{mIoU} & $\textbf{S}_{\alpha}$ & $\textbf{F}^w_{\beta}$ & $\textbf{mE}_{\phi}$ & \textbf{MAE} \\
\hline
P$_{only}$ & 0.789 & 0.879 & 0.839 & 0.937 & 0.026 \\
N$_{only}$ & 0.782 & 0.876 & 0.829 & 0.932 & 0.027 \\
Both     & 0.797 & 0.888 & 0.845 & 0.938 & 0.024 \\
\hline
\end{tabular}
}
\label{point_positive}
\end{table}
\begin{table}[h!]
\centering
\caption{Performance with different $\tau$ values.}
\resizebox{0.4\textwidth}{!}
{
\begin{tabular}{l|c|c|c|c|c}
\hline
\textbf{$\tau$} & \textbf{mIoU} & $\textbf{S}_{\alpha}$ & $\textbf{F}^w_{\beta}$ & $\textbf{mE}_{\phi}$ & \textbf{MAE} \\
\hline
0.3 & 0.794 & 0.882 & 0.842 & 0.938 & 0.025 \\
0.5 & 0.797 & 0.888 & 0.845 & 0.938 & 0.024 \\
0.7 & 0.786 & 0.877 & 0.832 & 0.933 & 0.027 \\
\hline
\end{tabular}
}
\label{hyper_settings}
\end{table}

\textbf{Effects of FPS.}
In the 4-5 rows of Tab.~\ref{ablation}, we validate the effectiveness of FPS.
Fig.~\ref{fig:fre_visual} intuitively shows sliding windows and prompt points obtained by FPS.
Guided by frequency domain prior information, the FPS effectively identifies point prompts, providing effective cues for refined segmentations.
In Tab.~\ref{FPS}, we investigate the effect of different hyperparameter settings.
It is important to note that the selected points consist of $k$ points with the highest values and $k$ points with the lowest, totaling 2$k$ points.
In Tab.~\ref{select_mode}, we investigate the effect of different point prompt methods.
The first row represents that we sample $2k$ points, randomly.
The second row represents that all the sliding windows across the entire image are selected, with each sliding window sized at 32 and $k$=2 points.
Obviously, points selected by FPS are more effective and the inference speed is much faster.
In Tab.~\ref{point_positive}, we investigate the impact of using only positive or only negative sample points.
The results demonstrate that incorporating both positive and negative sample points enables the model to generate more effective prompts.
Both positive and negative sample points located at boundaries are most helpful to segment animals from the background, respectively.
Moreover, we investigate the threshold for selecting positive and negative prompt points in the FPS.
As shown in Tab.~\ref{hyper_settings}, the results demonstrate that our method exhibits strong robustness.

\begin{figure}[!h]
\centering
\resizebox{0.46\textwidth}{!}
{
\renewcommand\arraystretch{0.1}
\begin{tabular}{@{}c@{}c}
\vspace{0.5mm}
\includegraphics[width=0.5\linewidth,height=2.5cm]{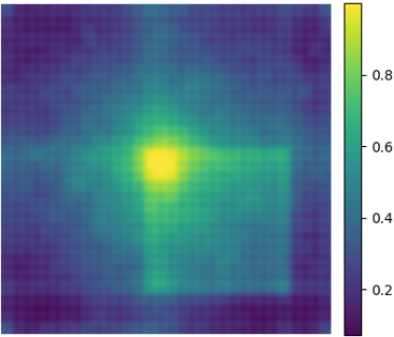}\ &
\includegraphics[width=0.5\linewidth,height=2.5cm]{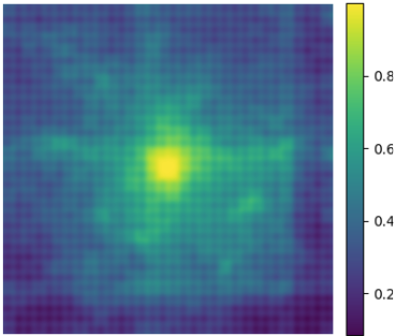}\  \\
\vspace{0.5mm}
  w/o FVM  &  w FVM  \\
\end{tabular}
}
\caption{Visualization of the effective receptive field brought by FVM.}
\label{fvm_erf}
\end{figure}

\begin{figure}[!h]
\centering
\resizebox{0.46\textwidth}{!}
{
\renewcommand\arraystretch{0.1}
\begin{tabular}{@{}c@{}c@{}c}
\vspace{0.5mm}
\includegraphics[width=0.33\linewidth,height=1.5cm]{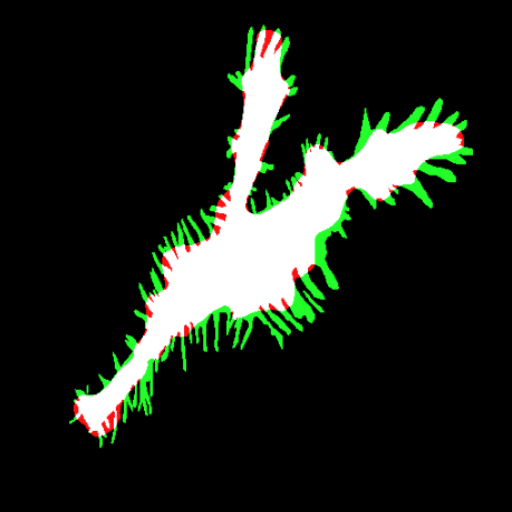}\ &
\includegraphics[width=0.33\linewidth,height=1.5cm]{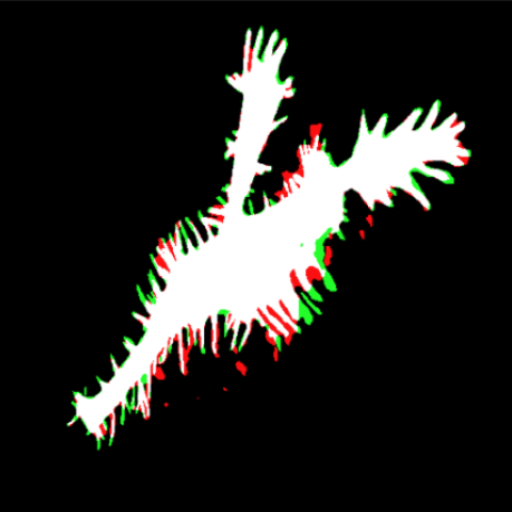}\ &
\includegraphics[width=0.33\linewidth,height=1.5cm]{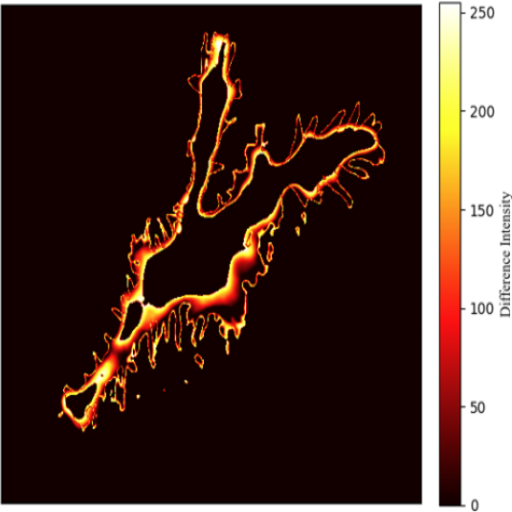}\ \\
\vspace{0.2mm}
\includegraphics[width=0.33\linewidth,height=1.5cm]{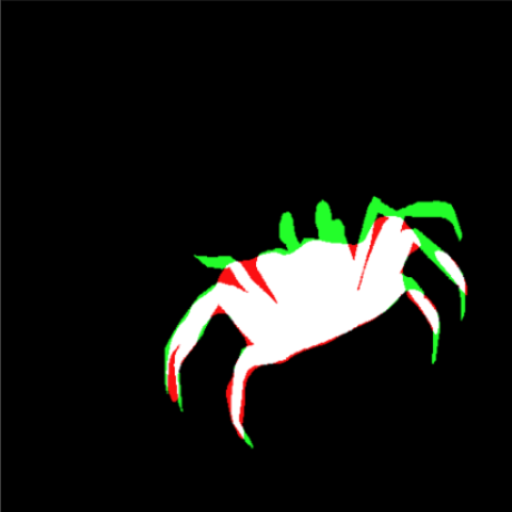}\ &
\includegraphics[width=0.33\linewidth,height=1.5cm]{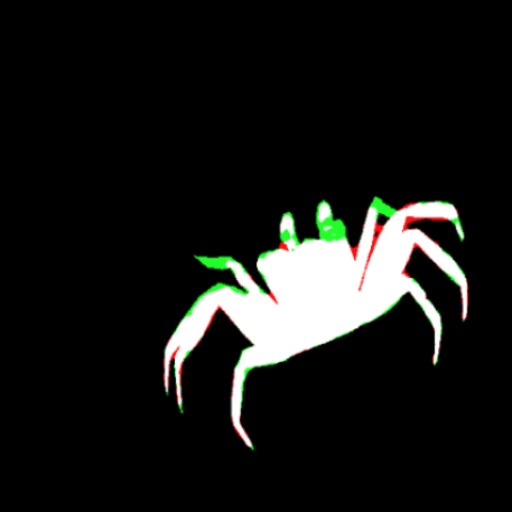}\ &
\includegraphics[width=0.33\linewidth,height=1.5cm]{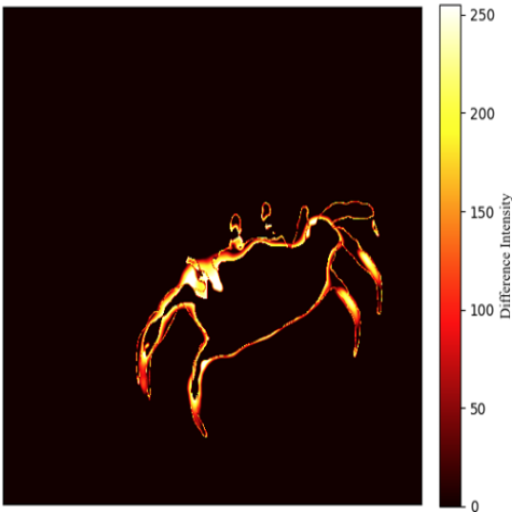}\ \\
\vspace{0.5mm}
 (a)  & (b) &(c)  \\
\end{tabular}
}
\caption{(a) and (b) are the segmentation results with and without using FVM against the ground truth, where green indicates correct predictions and red indicates redundant predictions. (c) is a heatmap showing the differences between the predictions in (a) and (b). Best view by zooming in.}
\label{fvm_edge}
\end{figure}

\textbf{Effects of FVM}
As shown in the 5-6 rows of Tab.~\ref{ablation}, the FVM can achieve a 2\% improvement in mIoU and show significant improvements across other metrics.
With the FVM, we fully utilize spatial and channel information from a comprehensive perspective.
In Fig.~\ref{fvm_erf}, we visually demonstrate the increase in the effective receptive field brought by the FVM.
In Fig.~\ref{fvm_edge}, we intuitively show how FVM aids in segmentation with complex boundaries.
We further investigate the effectiveness of each component within the FVM in Tab.~\ref{FVM}.
We compare our FVM with the Visual SSM (VSSM)~\cite{liu2024vmamba}, which selects features solely based on the height and width dimensions.
In the 2-7 rows of Tab.~\ref{FVM}, the experimental results demonstrate that both forward scanning and backward scanning can lead to performance improvements.

\textbf{Effect of Input Prompts.}
In our work, the prompt encoder can take in both points and coarse masks.
Tab.~\ref{tab:prompt_combo} shows the performance with different input prompts.
It consistently performs best by combining points with the coarse mask.
These results verify the complementarity between point-level boundary cues and the mask-level shape priors.
\begin{table}[h!]
\centering
\caption{Performance with different SSMs in FVM.}
\resizebox{0.44\textwidth}{!}
{
\begin{tabular}{l|c|c|c|c|c}
\hline
\textbf{Method} & \textbf{mIoU} & $\textbf{S}_{\alpha}$ & $\textbf{F}^w_{\beta}$ & $\textbf{mE}_{\phi}$ & \textbf{MAE} \\
\hline
Channel SSM & 0.771 & 0.868 & 0.823 & 0.924 & 0.029 \\
Spatial SSM & 0.775 & 0.872 & 0.825 & 0.926 & 0.029 \\
VSSM & 0.778 & 0.872 & 0.826 & 0.926 & 0.029 \\
FC-SSM & 0.790 & 0.885 & 0.838 & 0.933 & 0.026 \\
BC-SSM  & 0.783 & 0.874 & 0.828 & 0.932 & 0.028 \\
FVM  & 0.797 & 0.888 & 0.845 & 0.938 & 0.024 \\
\hline
\end{tabular}
}
\label{FVM}
\end{table}
\begin{table}[h!]
\centering
\caption{Performance with different input prompts.}
\resizebox{0.46\textwidth}{!}{
\begin{tabular}{l|c|c|c|c|c}
\hline
\textbf{Prompt} & \textbf{mIoU} & $\textbf{S}_{\alpha}$ & $\textbf{F}^w_{\beta}$ & $\textbf{mE}_{\phi}$ & \textbf{MAE} \\
\hline
Point & 0.767 & 0.870 & 0.825 & 0.926 & 0.028 \\
Mask & 0.759 & 0.866 & 0.822 & 0.925 & 0.029 \\
Point+Mask & 0.771 & 0.872 & 0.827 & 0.929 & 0.028 \\
\hline
\end{tabular}}
\label{tab:prompt_combo}
\end{table}

\begin{figure}[!t]
\centering
\includegraphics[width=0.48\textwidth]{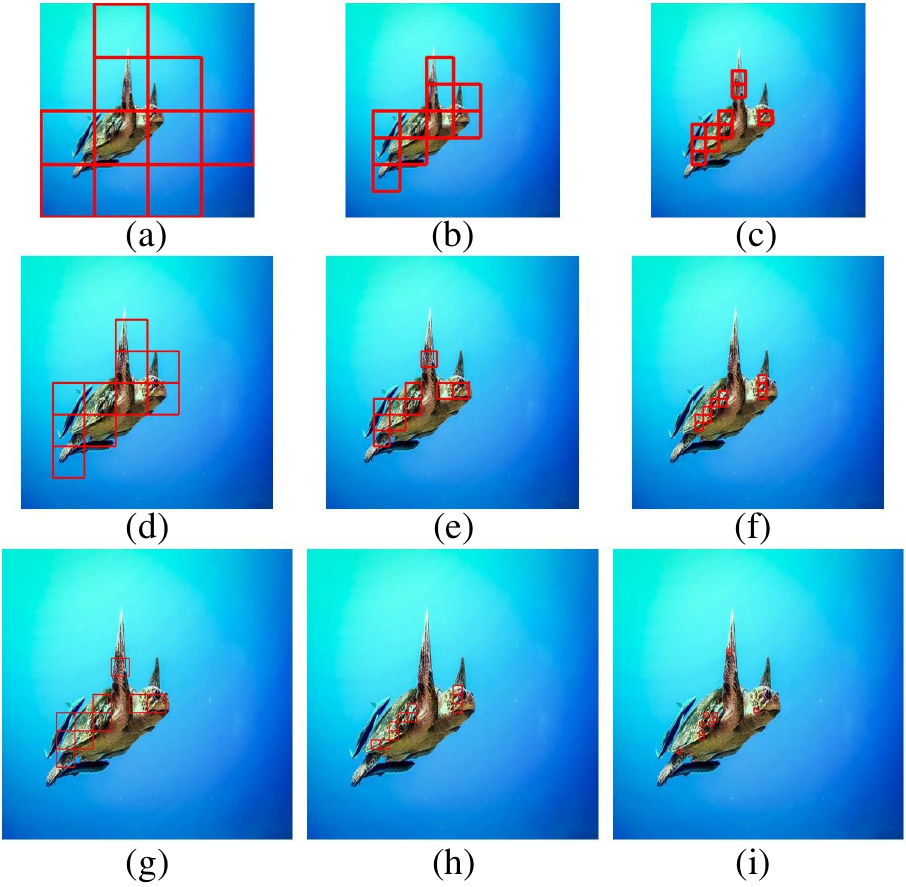}
\caption{Window selection in FPS. The image resolution increases from top to bottom. The window size decreases from left to right. Red rectangles show the selected top-\(k\) windows with \(k=10\).}
\label{fig:res_grid}
\end{figure}

\begin{figure}[!t]
\centering
{\setlength{\tabcolsep}{1pt}
\begin{tabular}{@{}c@{}c@{}}
\includegraphics[width=0.24\textwidth]{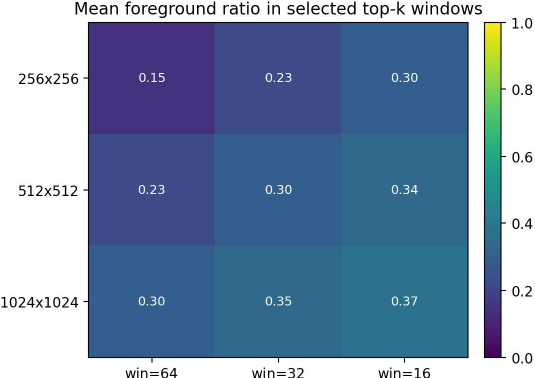} &
\includegraphics[width=0.24\textwidth,height=0.172\textwidth]{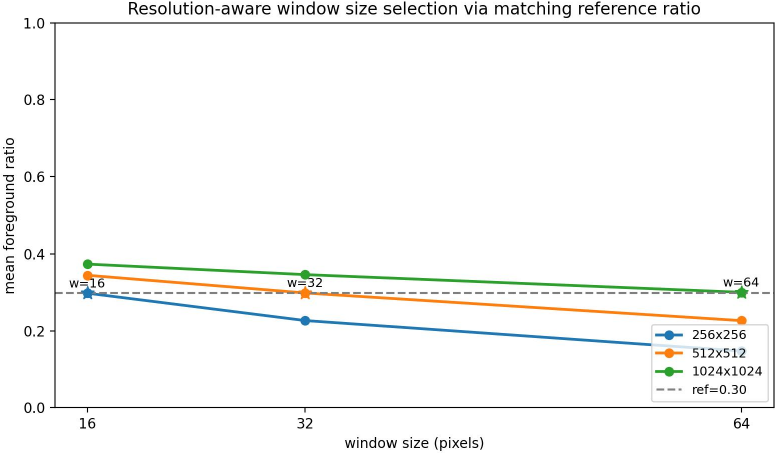} \\
\end{tabular}}
\caption{Resolution-aware window size selection with \(k=10\). Left: The mean foreground ratio. Right: Selected window sizes by matching the reference ratio from the default setting.}
\label{fig:posfrac_adaptive}
\end{figure}

\textbf{Effect of Image Resolutions for Window Sizes in FPS.}
The window size \(w\) is an important hyper-parameter in FPS.
The best window size may change with the image resolution.
To explore the effect, we use a fixed number of selected windows \(k=10\).
Meanwhile, we treat the default setting (resolution: \(512\times 512\), window size: 32) as our reference.
For different resolutions, we select the top-\(k\) windows and compute the mean foreground ratio in these windows.
Under the reference setting, the mean foreground ratio is around 0.30.
It indicates that many selected windows highlight object boundaries, and support sampling both positive and negative points.
This observation is consistent with Tab.~\ref{point_positive}.
Fig.~\ref{fig:res_grid} shows selected windows.
Fig.~\ref{fig:res_grid}(a)-(c) show the result with the setting of (resolution: 256$\times$ 256, window size: 64, 32, 16).
Fig.~\ref{fig:res_grid}(d)-(f) show the result with the setting of (resolution: 512$\times$ 512, window size: 64, 32, 16).
Fig.~\ref{fig:res_grid}(g)-(i) show the result with the setting of (resolution: 1024$\times$ 1024, window size: 64, 32, 16).
Fig.~\ref{fig:posfrac_adaptive} shows the mean foreground ratio statistics and curves.
It suggests that our method is robust to the window size.
As a result, we suggest \(w=16\) for 256$\times$ 256, \(w=32\) for 512$\times$ 512, and \(w=64\) for 1024$\times$ 1024.

\textbf{Effect of the Loss Weight $\lambda$.}
We further examine the Effect of the Loss Weight $\lambda$.
As shown in Tab.~\ref{tab:loss_weight}, varying \(\lambda\) from 0.5 to 2.0 results in only minor performance changes.
It indicates that our method is robust to this weight choice and supports the default \(\lambda=1\).
These results suggest that it provides a stable balance between the boundary refinement (BCE) and region-level overlap (IoU).
\begin{table}[h]
\centering
\caption{Effect of the loss weight on MAS3K.}
\label{tab:loss_weight}
\resizebox{0.4\textwidth}{!}{%
\begin{tabular}{c|c|c|c|c|c}
\hline
$\lambda$ & \textbf{mIoU} & $\textbf{S}_{\alpha}$ & $\textbf{F}^w_{\beta}$ & $\textbf{mE}_{\phi}$ & \textbf{MAE} \\
\hline
0.5 & 0.790 & 0.884 & 0.838 & 0.935 & 0.026 \\
1   & 0.797 & 0.888 & 0.845 & 0.938 & 0.024 \\
2   & 0.789 & 0.886 & 0.838 & 0.932 & 0.025 \\
\hline
\end{tabular}}
\end{table}
\subsection{Robustness to Synthetic Noise and Error Propagation}
To verify the noise resilience of frequency priors, we conduct a synthetic noise stress test.
We add Gaussian noise and speckle noise to the input images.
$\sigma$ is the noise strength (standard deviation).
As shown in Tab.~\ref{tab:noise_mas3k}, SAM+FGA consistently outperforms SAM across noise types and levels.
It supports that the frequency prior in FGA provides a robust guidance rather than learning dataset-specific noise templates.
\begin{table}[h!]
\centering
\caption{Noise stress test on MAS3K. mIoU is reported.}
\resizebox{0.4\textwidth}{!}
{
\begin{tabular}{c|cc|cc}
\hline
& \multicolumn{2}{c|}{\textbf{Gaussian}} & \multicolumn{2}{c}{\textbf{Speckle}} \\
\cline{2-5}
\textbf{$\sigma$} & \textbf{SAM} & \textbf{SAM+FGA} & \textbf{SAM} & \textbf{SAM+FGA} \\
\hline
0.00 & 0.566 & 0.754 & 0.566 & 0.754 \\
0.05 & 0.546 & 0.670 & 0.549 & 0.708 \\
0.10 & 0.529 & 0.663 & 0.532 & 0.683 \\
0.20 & 0.518 & 0.598 & 0.526 & 0.665 \\
\hline
\end{tabular}
}
\label{tab:noise_mas3k}
\end{table}

When the coarse mask is unreliable, sampled points from low-confidence windows may introduce noises and further amplify errors.
To mitigate this issue, we compute a window confidence score based on the coarse mask.
If too few windows remain after filtering, we fall back to using the coarse mask only, which prevents the prompt set from being dominated by unreliable samples.
Tab.~\ref{tab:gated_fps} provides an error analysis to quantify the failure frequency from two perspectives: grid-level prompt errors (low-IoU selected windows) and point-level prompt errors.
As observed, the confidence gating reduces both grid-level failure frequency (\emph{e.g.}, the grid-level prompt error drops from 9.01\% to 6.92\%) and point-level prompt errors (\emph{e.g.}, the point-level prompt error drops from 5.05\% to 2.79\%), demonstrating a feasible mitigation effect.
\begin{table}[h!]
\centering
\caption{Error analysis of confidence-gated FPS.}
\resizebox{0.46\textwidth}{!}{
\begin{tabular}{l|c|c}
\hline
\textbf{Metric} & \textbf{FPS} & \textbf{FPS + gating (\(\gamma=0.8\))} \\
\hline
\multicolumn{3}{c}{\textbf{Grid-level}} \\
\hline
Grid-level error & 9.01\% & \textbf{6.92\%} \\
Image-level: at least one failing grid & 35.41\% & \textbf{29.01\%} \\
Image-level: failing grids \(\ge 2\) & 20.86\% & \textbf{13.67\%} \\
Image-level: failing grids \(\ge 50\%\) & 5.00\% & \textbf{4.38\%} \\
\hline
\multicolumn{3}{c}{\textbf{Point-level}} \\
\hline
Point-level error & 5.05\% & \textbf{2.79\%} \\
Positive point error (FP) & 16.78\% & \textbf{11.84\%} \\
Negative point error (FN) & 0.88\% & \textbf{0.67\%} \\
Image-level: at least one wrong point & 35.76\% & \textbf{18.84\%} \\
Image-level: error rate \(\ge 20\%\) & 8.41\% & \textbf{5.17\%} \\
\hline
\end{tabular}}
\label{tab:gated_fps}
\end{table}
\begin{figure}[!h]
\centering
\resizebox{0.48\textwidth}{!}{
\begin{tabular}{@{}c@{}c@{}c}
\vspace{0.5mm}
\includegraphics[width=0.33\linewidth,height=1.5cm]{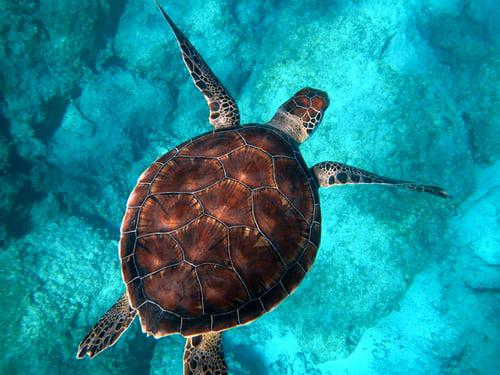}\ &
\includegraphics[width=0.33\linewidth,height=1.5cm]{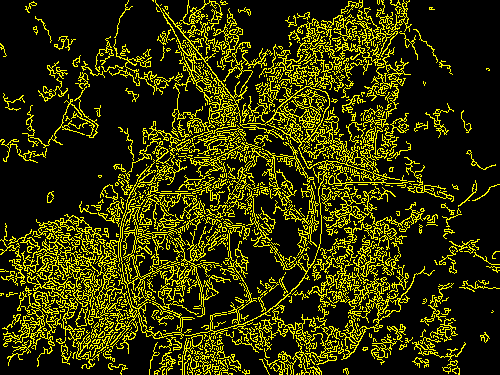}\ &
\includegraphics[width=0.33\linewidth,height=1.5cm]{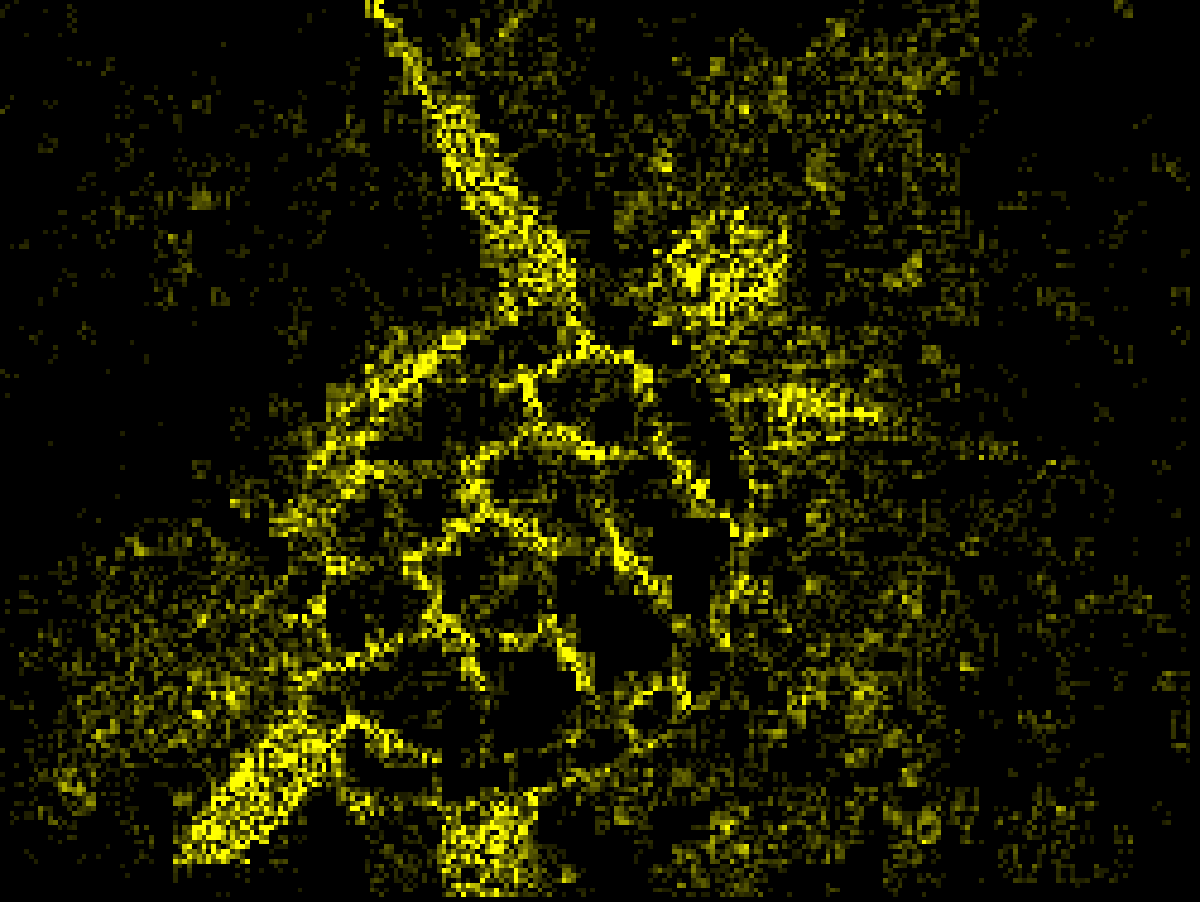}\ \\
\vspace{0.5mm}
 (a)  & (b) &(c)  \\
\end{tabular}}
\caption{Visual effect of our frequency domain priors. (a) Input Images; (b) The result by using canny edge detection; (c) The result by our frequency domain method. Best view by zooming in.}
\label{canny_fre}
\end{figure}
\subsection{Distinction Between Frequency Information and Edge}
Our task involves image segmentation in marine environments, which are notably complex.
Traditional edge detection methods like Canny~\cite{canny2009computational} are sensitive to noise.
Although using preliminary smoothing, edge detection cannot fully mitigate the impact of residual noise on the effective edge.
In contrast, our work extracts frequency domain features using wavelet transforms.
Wavelet transforms use multi-scale analysis and process images at different scales, separating features and noise of various scales.
Subsequently, the soft-thresholding operation removes noise in the high-frequency components while preserving edge information in the low-frequency components.
As shown in Fig.~\ref{canny_fre}, when faced with situations where the background noise is extremely high, edge detection can hardly obtain effective information.
However, the frequency domain prior information obtained after wavelet transformations still provides effective cues.
\subsection{Computational Efficiency}
To evaluate the computational efficiency, we report the end-to-end latency (ms/frame) and peak GPU memory (MB) in \ref{cost}.
More specifically, we provide a runtime breakdown for the SAM' image encoder (En), SAM' mask encoder (De1), FGA, FPS, SAM' learnable mask encoder (De2) and FVM.
The latency is measured on the same RTX 3090 GPU with CUDA synchronization, using 30 warmup runs and averaging over 200 times.
The peak memory is reported as the maximum allocated GPU memory during inference.
For a fair comparison, all variants share the same input resolution and precision setting.
The results show that our method is rather efficient.
\begin{table}[h!]
\centering
\caption{Comparison of the latency and memory cost.}
\resizebox{0.5\textwidth}{!}{%
\begin{tabular}{l|c|c|c|c|c|c|c|c}
\hline
\textbf{Variant} & \textbf{En} & \textbf{De1} & \textbf{FGA} & \textbf{FPS} & \textbf{De2} & \textbf{FVM} & \textbf{Overall} & \textbf{Peak Mem}\\
\hline
SAM         & 21.04 & 15.23 & 0.00 & 0.00 & 0.00 & 0.00 & 36.38 & 618\\
SAM+FGA     & 27.06 & 15.00 & 0.54 & 0.00 & 0.00 & 0.00 & 42.71 & 649\\
SAM+FGA+FPS & 27.06 & 15.13 & 0.45 & 1.18 & 15.65 & 0.00 & 59.77 & 654\\
HFP-SAM     & 27.06 & 15.25 & 0.44 & 1.19 & 15.80 & 0.15 & 59.92 & 654\\
\hline
\end{tabular}}
\label{cost}
\end{table}
\begin{table}
\centering
\caption{Performance comparison on COD task.}
\resizebox{0.4\textwidth}{!}{
\begin{tabular}{c|c|c|c|c}
\hline
&\multicolumn{4}{c}{\textbf{COD10K~\cite{fan2020camouflaged}}}  \\ \cline{2-5}
\textbf{Method} & $\textbf{S}_\alpha$ & $\textbf{F}_\beta^w$ & $\textbf{m}\textbf{E}_\phi$ & \textbf{MAE} \\
\hline
SINet~\cite{fan2020camouflaged} & 0.771 & 0.551 & 0.806 & 0.051 \\
RankNet~\cite{lv2021simultaneously} & 0.767 & 0.611 & 0.861 & 0.045 \\
PFNet~\cite{mei2021camouflaged}  & 0.800 &0.660&0.868&0.040\\
FBNet~\cite{lin2023frequency} & 0.809 &0.684&0.889&0.035\\
ECDNet~\cite{li2021marine}  & 0.783 &0.701&0.798&0.050\\
BCMNet~\cite{cheng2023bidirectional} & 0.829 &0.723&0.899&0.027\\
\hline
HFP-SAM & 0.912 &0.891&0.982&0.011\\
\hline
\end{tabular}}
\label{cod}
\end{table}
\subsection{Model Transferability}
To verify the generalization capability of our method, we conduct experiments on a diverse set of vision tasks, including camouflaged object detection~\cite{fan2020camouflaged}, polyp segmentation~\cite{jha2020kvasir}, and salient object detection~\cite{wang2017learning}.
These tasks represent a wide range of segmentation scenarios.
Camouflaged object detection focuses on identifying subtle and hidden objects within complex backgrounds.
Polyp segmentation addresses medical segmentation requirements, targeting small and irregular medical features.
Salient object detection emphasizes identifying visually prominent objects in natural scenes.
In Tab.~\ref{cod}, Tab.~\ref{kva} and Tab.~\ref{dut}, we present a performance comparison of our method against relevant approaches in corresponding tasks.
The results demonstrate a clear performance advantage of our method over relevant approaches in these tasks.
This highlights the versatility of our method across varied segmentation contexts.
It also demonstrates its strong generalization ability, achieving state-of-the-art performance in scenarios with distinct characteristics and challenges.
\begin{table}
\centering
\caption{Performance comparison on PS task.}
\resizebox{0.4\textwidth}{!}
{
\begin{tabular}{c|c|c|c|c}
\hline
&\multicolumn{4}{c}{\textbf{Kvasir~\cite{jha2020kvasir}}}  \\ \cline{2-5}
\textbf{Method} & $\textbf{S}_\alpha$ & $\textbf{F}_\beta^w$ & $\textbf{m}\textbf{E}_\phi$ & \textbf{MAE} \\
\hline
SANet~\cite{ren2020salient} & 0.915 & 0.892 & 0.953 & 0.028 \\
C2FNet~\cite{zhuge2022salient}& 0.905 &0.870&0.889&0.035 \\
LDNet~\cite{liu2021visual}  & 0.905&0.869&0.945&0.031\\
FAPNet~\cite{wu2022edn}  & 0.919 & 0.894 & 0.953 & 0.027 \\
CFANet~\cite{wang2023pixels}  & 0.924 & 0.903 & 0.962 & 0.023 \\
\hline
HFP-SAM & 0.925 &0.909&0.964&0.023\\
\hline
\end{tabular}
}
\label{kva}
\end{table}
\begin{table}\centering
\caption{Performance comparison on SOD task.}
\resizebox{0.4\textwidth}{!}
{
\begin{tabular}{c|c|c|c|c}
\hline
&\multicolumn{4}{c}{\textbf{DUTS~\cite{wang2017learning}}}  \\ \cline{2-5}
\textbf{Method} & $\textbf{S}_\alpha$ & $\textbf{F}_\beta^w$ & $\textbf{m}\textbf{E}_\phi$ & \textbf{MAE} \\
\hline
CANet~\cite{ren2020salient} & 0.878 & 0.838 & 0.889 & 0.044 \\
ICON~\cite{zhuge2022salient}& 0.809 &0.684&0.889&0.035 \\
VST~\cite{liu2021visual}  & 0.896 &0.818&0.892&0.037\\
EDN~\cite{wu2022edn}  & 0.883 & 0.845 & 0.911 & 0.041 \\
MENet~\cite{wang2023pixels}  & 0.905 & 0.893 & 0.937 & 0.028 \\
BBRF~\cite{ma2023boosting}  & 0.908 & 0.893 & 0.927 & 0.025 \\
\hline
HFP-SAM & 0.915 &0.899&0.951&0.025\\
\hline
\end{tabular}
}
\label{dut}
\end{table}
\subsection{Zero-shot Experiments}
To verify the zero-shot ability, we conduct experiments on datasets of camouflaged object detection, polyp segmentation and salient object detection, i.e., COD10K~\cite{fan2020camouflaged}, Kvasir~\cite{jha2020kvasir} and DUTS~\cite{wang2017learning}.
To this end, we directly evaluate SAM and our proposed HFP-SAM on these tasks without any fine-tuning.
As shown in Tab.~\ref{zero-shot1}, our method consistently achieves better results to the original SAM.
This confirms that our method retains SAM's zero-shot capability.
\begin{table}
\centering
\caption{Performance comparison of zero-shot capability.}
\resizebox{0.40\textwidth}{!}
{
\begin{tabular}{c|c|c|c|c}
\hline
&\multicolumn{4}{c}{\textbf{COD10K}}  \\ \cline{2-5}
\textbf{Method} & $\textbf{S}_\alpha$ & $\textbf{F}_\beta^w$ & $\textbf{m}\textbf{E}_\phi$ & \textbf{MAE} \\
\hline
SAM~\cite{kirillov2023segment} & 0.765 & 0.633 & 0.835 & 0.050 \\
\hline
HFP-SAM & 0.786 &0.668&0.853&0.046\\
\hline
&\multicolumn{4}{c}{\textbf{Kvasir}}  \\ \cline{2-5}
\textbf{Method} & $\textbf{S}_\alpha$ & $\textbf{F}_\beta^w$ & $\textbf{m}\textbf{E}_\phi$ & \textbf{MAE} \\
\hline
SAM~\cite{kirillov2023segment} & 0.356 & 0.246 & 0.339 & 0.515 \\
\hline
HFM-SAM & 0.392 &0.270&0.391&0.464\\
\hline
&\multicolumn{4}{c}{\textbf{DUTS}}  \\ \cline{2-5}
\textbf{Method} & $\textbf{S}_\alpha$ & $\textbf{F}_\beta^w$ & $\textbf{m}\textbf{E}_\phi$ & \textbf{MAE} \\
\hline
SAM~\cite{kirillov2023segment} & 0.831 & 0.764 & 0.879 & 0.058 \\
\hline
HFP-SAM & 0.839 &0.793&0.893&0.053\\
\hline
\end{tabular}
}
\label{zero-shot1}
\end{table}
\subsection{Failure Cases and Limitations}
\begin{figure}[!t]
\centering
\resizebox{0.5\textwidth}{!}
{
\renewcommand\arraystretch{0.1}
\begin{tabular}{@{}c@{}c@{}c@{}c}
\vspace{0.5mm}
\includegraphics[width=0.245\linewidth,height=2.0cm]{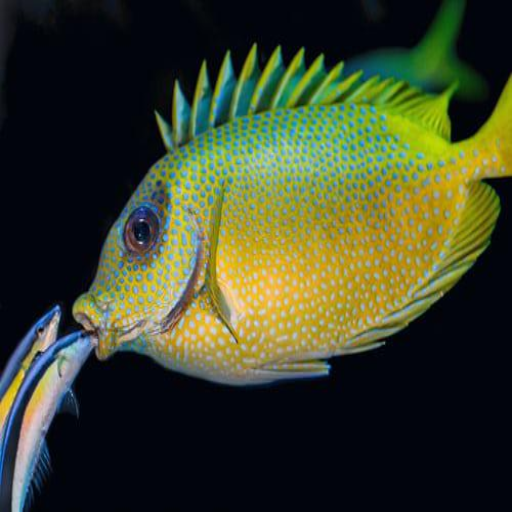}\ &
\includegraphics[width=0.245\linewidth,height=2.0cm]{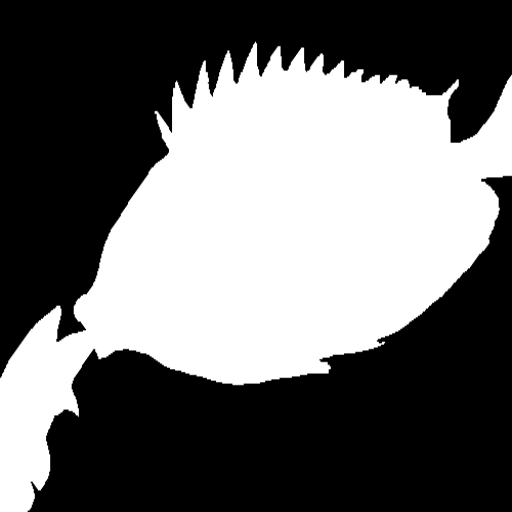}\ &
\includegraphics[width=0.245\linewidth,height=2.0cm]{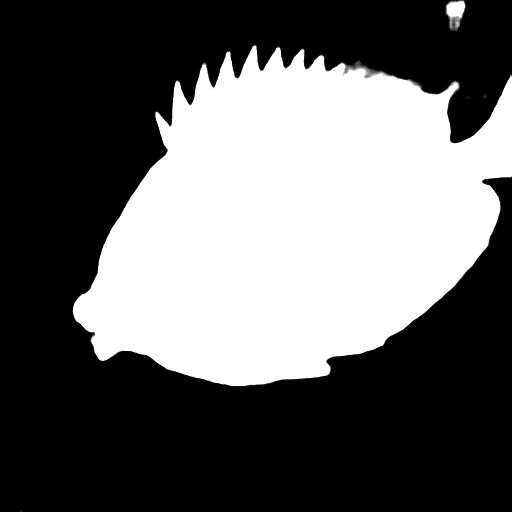}\ &
\includegraphics[width=0.245\linewidth,height=2.0cm]{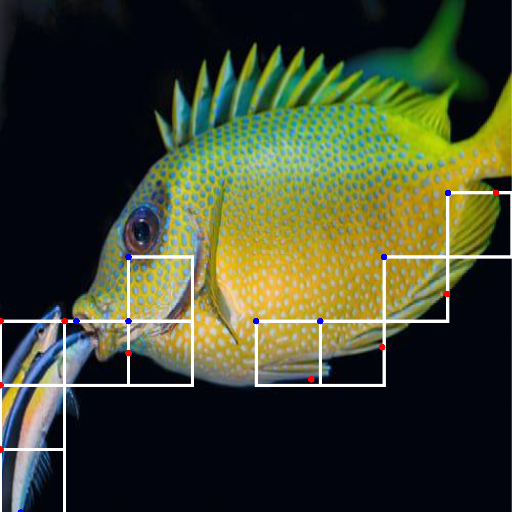}\ \\
\vspace{0.5mm}
\includegraphics[width=0.245\linewidth,height=2.0cm]{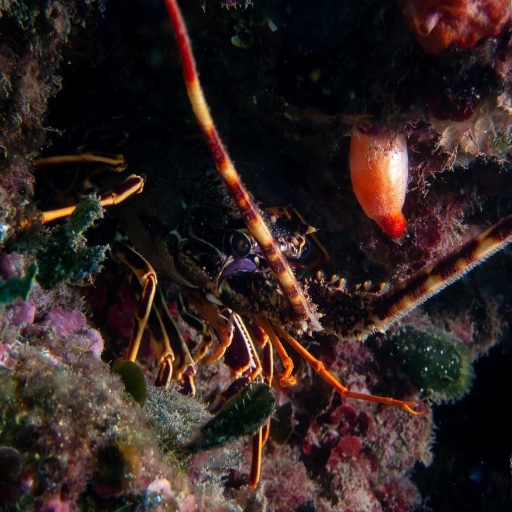}\ &
\includegraphics[width=0.245\linewidth,height=2.0cm]{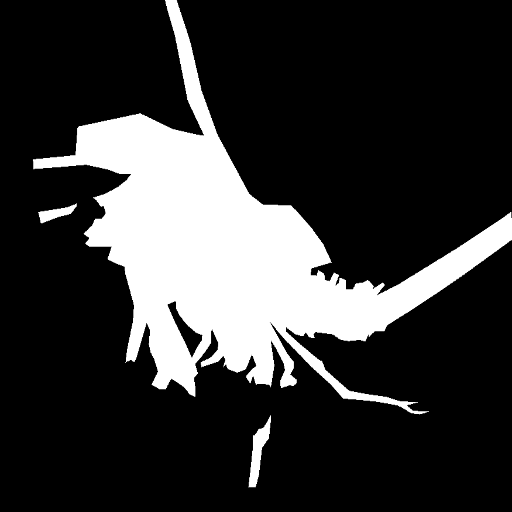}\ &
\includegraphics[width=0.245\linewidth,height=2.0cm]{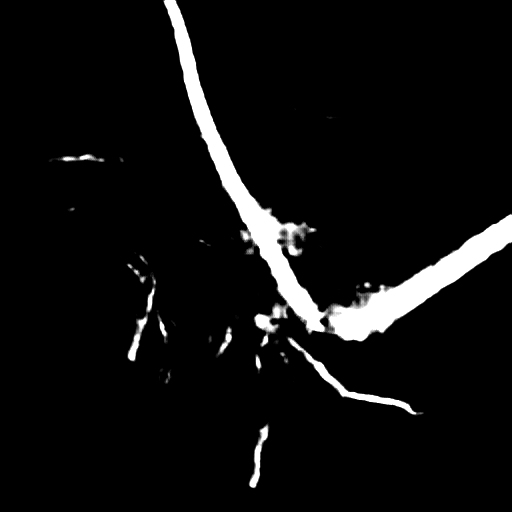}\ &
\includegraphics[width=0.245\linewidth,height=2.0cm]{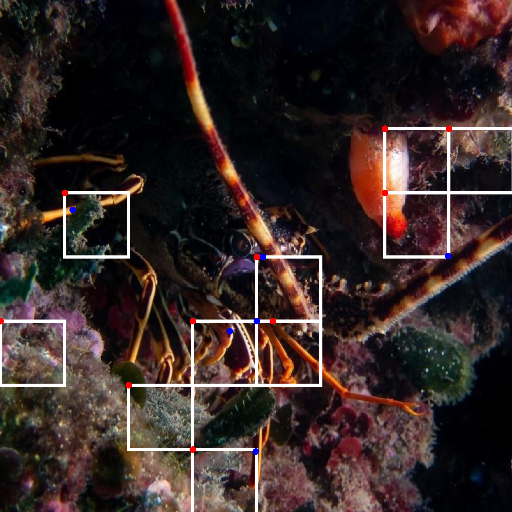}\ \\
\vspace{0.5mm}
Image & GT & Coarse Mask & Point Prompts \\
\end{tabular}
}
\caption{Visualizations of failure cases and the corresponding point prompts obtained by our FPS. Best view by zooming in.}
\label{fail}
\end{figure}

If significant errors appear in the coarse segmentation, it can lead to the issue of amplifying these mistakes in the process of the fine segmentation.
As shown in Fig.~\ref{fail}, the coarse segmentation mask overlooks small fish.
Consequently, the point prompts obtained from our FPS may be wrong.
Thus, our model can not refine segmentation results during the second phase.
In the future, we will explore more effective techniques to address this limitation.
\section{Conclusion}
In this paper, we propose a novel framework named HFP-SAM for high performance MAS.
Specifically, we first introduce a Frequency Guided Adapter (FGA) to integrate frequency domain priors into SAM.
Then, we propose a Frequency-aware Point Selection (FPS) to generate efficient point prompts for SAM.
Furthermore, we design a Full View Mamba (FVM) to select features from both spatial and channel views.
With FGA and FPS, HFP-SAM leverages frequency information to enhance SAM's representation ability.
With FVM, it achieves precise segmentation masks by fully integrating frequency and spatial domain information.
Experiments on four MAS datasets validate the effectiveness of our method.
\bibliographystyle{IEEEtran}
\bibliography{IEEEabrv,main}
\end{document}